\documentclass[times,twocolumn,final]{elsarticle}
\usepackage{medima}
\usepackage{framed,multirow}

\usepackage{amsmath,amssymb}
\usepackage{latexsym}

\usepackage{url}
\usepackage{xcolor}
\usepackage{adjustbox}
\usepackage{threeparttable}
\usepackage{booktabs}
\usepackage{multirow}
\usepackage{tabularx}

\usepackage{algorithm}
\usepackage{algorithmic}

\newcommand{\mb}[1]{\mathbf{#1}}

\definecolor{airforceblue}{rgb}{0.36, 0.54, 0.66}
\definecolor{ballblue}{rgb}{0.004, 0.50, 0.69}
\definecolor{newcolor}{rgb}{.8,.349,.1}

\usepackage[colorlinks]{hyperref}

\usepackage[utf8]{inputenc}
\usepackage{graphicx}

\AtBeginDocument{%
  \hypersetup{
    citecolor=ballblue,
    linkcolor=ballblue,   
    urlcolor=ballblue}}

\journal{Medical Image Analysis}

\begin{document}

\verso{Xiaofei Wang \textit{et~al.}}

\begin{frontmatter}
\title{Joint Modelling Histology and Molecular Markers for Cancer Classification\tnoteref{tnote1}}

\author[1]{Xiaofei Wang\fnref{fn1}}
\author[2]{Hanyu Liu\fnref{fn1}}
\fntext[fn1]{Equal contribution.}
\author[1]{Yupei Zhang}
\author[2]{Boyang Zhao}
\author[3]{Hao Duan}
\author[4]{Wanming Hu}
\author[3]{Yonggao Mou}
\author[1]{Stephen Price}
\author[1,2,5,6]{Chao Li\corref{cor1}}
\cortext[cor1]{Corresponding author.}\ead{cl647@cam.ac.uk}

\address[1]{Department of Clinical Neurosciences, University of Cambridge, UK}
\address[2]{School of Science and Engineering, University of Dundee, UK}
\address[3]{Department of Neurosurgery, State Key Laboratory of Oncology in South China, Guangdong Provincial Clinical Research Center for Cancer, Sun Yat-sen University Cancer Center, China}
\address[4]{Department of Pathology, State Key Laboratory of Oncology in South China, Guangdong Provincial Clinical Research Center for Cancer, Sun Yat-sen University Cancer Center, China}
\address[5]{Department of Applied Mathematics and Theoretical Physics, University of Cambridge, UK}
\address[6]{School of Medicine, University of Dundee, UK}
% ...

\received{-}
\finalform{-}
\accepted{-}
\availableonline{-}
\communicated{-}

%%%
\begin{abstract}

Cancers are characterized by remarkable heterogeneity and diverse prognosis.
Accurate cancer classification is essential for patient stratification and clinical decision-making.
 Although digital pathology has been advancing cancer diagnosis and prognosis, the paradigm in cancer pathology has shifted from purely relying on histology features to incorporating molecular markers.
There is an urgent need for digital pathology methods to meet the needs of the new paradigm. 
We introduce a novel digital pathology approach to jointly predict molecular markers and histology features and model their interactions for cancer classification. Firstly, 
%, in order to extract efficient embeddings for both genotype and phenotype prediction, 
to mitigate the challenge of cross-magnification information propagation, we propose a multi-scale disentangling module, enabling the extraction of multi-scale features from high-magnification (cellular-level) to low-magnification (tissue-level) whole slide images.
Further, based on the multi-scale features, we propose an attention-based hierarchical multi-task multi-instance learning framework to simultaneously predict histology and molecular markers. Moreover, we propose a co-occurrence probability-based label correlation graph network to model the co-occurrence of molecular markers.
Lastly, we design a cross-modal interaction module with the dynamic confidence constrain loss and a cross-modal gradient modulation strategy, to model the interactions of histology and molecular markers. 
Our experiments demonstrate that our method outperforms other state-of-the-art methods in classifying glioma, histology features and molecular markers. Our method promises to promote 
precise oncology with the potential to advance biomedical research and clinical applications. The code is available at \href{https://github.com/LHY1007/M3C2}{github}.

\end{abstract}


\end{frontmatter}

%\linenumbers

%% main text

\section{Introduction}

Cancers are malignant tumors that are lethal to humans. Early and accurate diagnosis is crucial for managing cancers. Currently, pathology remains the gold standard for diagnosing cancers. However,  traditional pathology is labour-intensive, time-consuming, and heavily dependent on the expertise of neuropathologists. Digital pathology, which utilizes automated algorithms to analyze  tissue whole slide images (WSIs) \citep{lu2021data}, holds the promise of rapid diagnosis for timely and precise treatment.

Recently, deep learning (DL)-based digital pathology approaches have been successfully applied in diagnosing various cancers \citep{jose2023artificial}. These methods primarily focus on cancer diagnosis using histology features according to pathology diagnostic criteria. Alongside these efforts, the past decade has witnessed a surge in the discovery of molecular markers for cancer diagnosis, catalyzing a paradigm shift in the cancer diagnostic criteria from traditional histopathology towards molecular pathology. For instance, glioma is one of the most prevalent malignant primary tumors in adults, with a median overall survival of less than 14 months for high-grade glioma \citep{molinaro2019genetic,zhang2024phy}.
The 2021 WHO Classification of Brain Tumours \citep{louis20212021} has established several key molecular markers, such as isocitrate dehydrogenase (IDH) mutations, co-deletion of chromosome 1p/19q, and homozygous deletion (HOMDEL) of cyclin-dependent kinase inhibitor 2A/B (CDKN). According to the updated criteria, glioblastoma is primarily diagnosed based on IDH mutations, whereas it was previously based purely on histology features such as necrosis and microvascular proliferation (NMP). In addition, a new class of high grade astrocytoma has been defined by IDH and CDKN. This prominent paradigm shift  has revolutionized the clinical pathway for cancers and offers new opportunities for developing novel digital pathology approaches.

Integrating molecular markers and histology into clinical diagnosis still faces several  practical challenges. Firstly, the methods of assessing molecular markers, such as gene sequencing and immunostaining, are often time-consuming and costly. Secondly, histology assessment and molecular markers follow separate diagnostic workflows, hindering timely and effective integration for clinical diagnosis. DL approaches promise to tackle the above challenges with  unique advantages: 1) Predicting molecular markers from WSIs: Mounting research indicates that histology features are associated with molecular alterations, making it feasible to predict molecular markers directly from WSIs \citep{wang2024cross}. 2)  Integrating diagnostic flows: DL-based approaches can effectively integrate the diagnostic flows of molecular markers and histology features. Additionally, they can model the interactions between data types, thereby advancing cancer diagnosis under the latest diagnostic paradigm.
There is a pressing need for developing DL approaches to jointly predict molecular markers and histology while capturing their interactions.This integrated approach holds the potential to streamline and enhance the accuracy of cancer diagnosis.

In this paper, we introduce a novel digital pathology approach, namely multi-scale multi-task modelling for cancer classification (M3C2), aligning with the emerging paradigm of molecular pathology for cancer diagnosis. 
Previous approaches have attempted to integrate histology and genomics for tumour diagnosis \citep{xing2022discrepancy}. For example, \cite{ding2023pathology}  developed a multi-modal transformer with  unsupervised pretraining to integrate pathology and genomics for predicting colon cancer survival.
Despite success, most existing methods  only use molecular markers as supplementary input. They are incapable of simultaneously predicting histology and molecular markers and further modelling their interaction, which limits their clinical relevance under current diagnostic scheme.
To address this limitation and align with the updated clinical diagnostic pathway, we leverage a novel hierarchical multi-task framework based on vision transformer \citep{dosovitskiy2020image}. Our approach features two partially weight-sharing components that jointly predict histology and molecular markers, effectively modeling their interactions and enhancing clinical relevance.

First, we design our model to extract informative features from multiple  WSI magnifications. In clinical practice, pathology diagnosis involves inspecting tissue sections under different magnification, e.g., from 20X (0.25 $\mu$m $\textrm{px}^{-1}$ at cellular-level) to 10X (1 $\mu$m $\textrm{px}^{-1}$ at tissue-level) \citep{schmitz2021multi}. Hence, we design a multi-scale disentangling module to capture the crucial WSI features required for cancer diagnosis.
Specifically, we employ a novel disentanglement loss to efficiently extract features for both histology and molecular markers.

Further, we focus on modelling relationship among different molecular markers. It is known that molecular markers are inherently associated due to underlying cancer evolution and biology. Consequently, multiple molecular markers are often required for accurate cancer classification according to guidelines. To mirror real-world scenarios, we formulate the prediction of multiple molecular markers as a multi-label classification (MLC) task. Although previous MLC methods have effectively captured label correlations \citep{li2022multi}, existing approaches may overlook the co-occurrence and intrinsic associations of molecular markers during prediction \citep{zhang2023impact}.  To address this gap, we propose a co-occurrence probability-based label-correlation graph (CPLC-Graph) network to model the co-occurrence and relationships of molecular markers.

Lastly, we model the interaction between molecular markers and histology predictions. Specifically, we introduce a  cross-modal interaction module to capture the interplay between molecular markers and histology features, such as IDH mutation and NMP, both crucial for diagnosing glioblastoma. In this module, we firstly design a dynamic confidence constrain (DCC) loss, guiding the model to focus on similar WSI regions for both tasks. Beyond loss-level interaction, we also develop the cross-modal gradient modulation (CMG-Modu) learning strategy to coordinate the training process of histology and molecular marker predictions.
To our best knowledge, this is the first attempt to classify cancer via modeling the interaction of histology and molecular markers predictions.

Our main contributions are summarized as follows. (1) We propose a multi-task multi-instance learning framework to jointly predict histology and molecular markers and classify glioma, reflecting the up-to-date diagnosis paradigm. (2) We devise a multi-scale disentangling module to generate efficient multi-scale features for both  histology and molecular marker prediction. A CPLC-Graph network is proposed to model the  relationship of multiple molecular markers.  
(3) We design a DCC loss and a CMG-Modu training strategy  to coordinate the cross-modal interaction between histology and molecular markers for glioma classification. 
(4) We perform extensive experiments to validate our performance in several tasks of glioma classification,
molecular markers and histology prediction.

\section{Related work}

In traditional pathology,  WSI inspection by pathologists remains the gold standard for clinical diagnosis \citep{deng2020deep}. However, the conventional diagnostic process is labor-intensive and time-consuming \citep{deng2020deep}, with the shortage of experienced pathologists further exacerbating the challenge. DL-based methods \citep{martinez2023computer}, which employ biologically-inspired neural networks to utilize the rich information contained in WSIs. are advancing computer-aided pathology diagnosis. In this section, we firstly review DL-based digital pathology methods for cancer diagnosis, and then focus on  methods specific to glioma diagnosis, and finally review methods that integrate histology and molecular markers for precision oncology.

\subsection{Digital pathology for cancer diagnosis}

DL-based digital pathology methods are generally  categorized into low (cell)-level modelling and high (tissue)-level modelling. These approaches address downstream tasks such as cell segmentation \citep{graham2019hover} and tissue phenotyping  \citep{bhattacharyya2024efficient}, respectively. Specifically, cell segmentation models primarily focus on assigning pixel-level labels for cells and nuclei \citep{graham2019hover}, further used as biomarkers for disease diagnosis. For instance, \cite{graham2019hover} proposed Hover-Net, which performs simultaneous nuclear segmentation and classification by leveraging the distances of nuclear pixels to the mass centre. Similarly, \cite{ren2021deep}  developed a DL-based pipeline to generate an embedded map profiling cell composition,  further utilized to extract texture patterns for tumour and immune cells. 
In contrast to cell-level modelling, tissue-level modelling is more efficient in capturing global phenotype features,  providing a more comprehensive understanding of tissue architecture and pathology that is crucial for accurate and effective clinical assessments. Particularly, recent DL-based approaches  have significantly improved computer-aided classification of various diseases \citep{li2022comprehensive}, including membranous nephropathy \citep{hao2023accurate}, interstitial lung disease \citep{uegami2022mixture}.

  \begin{figure*}[t]
    \centering
    \includegraphics[width=1\linewidth]{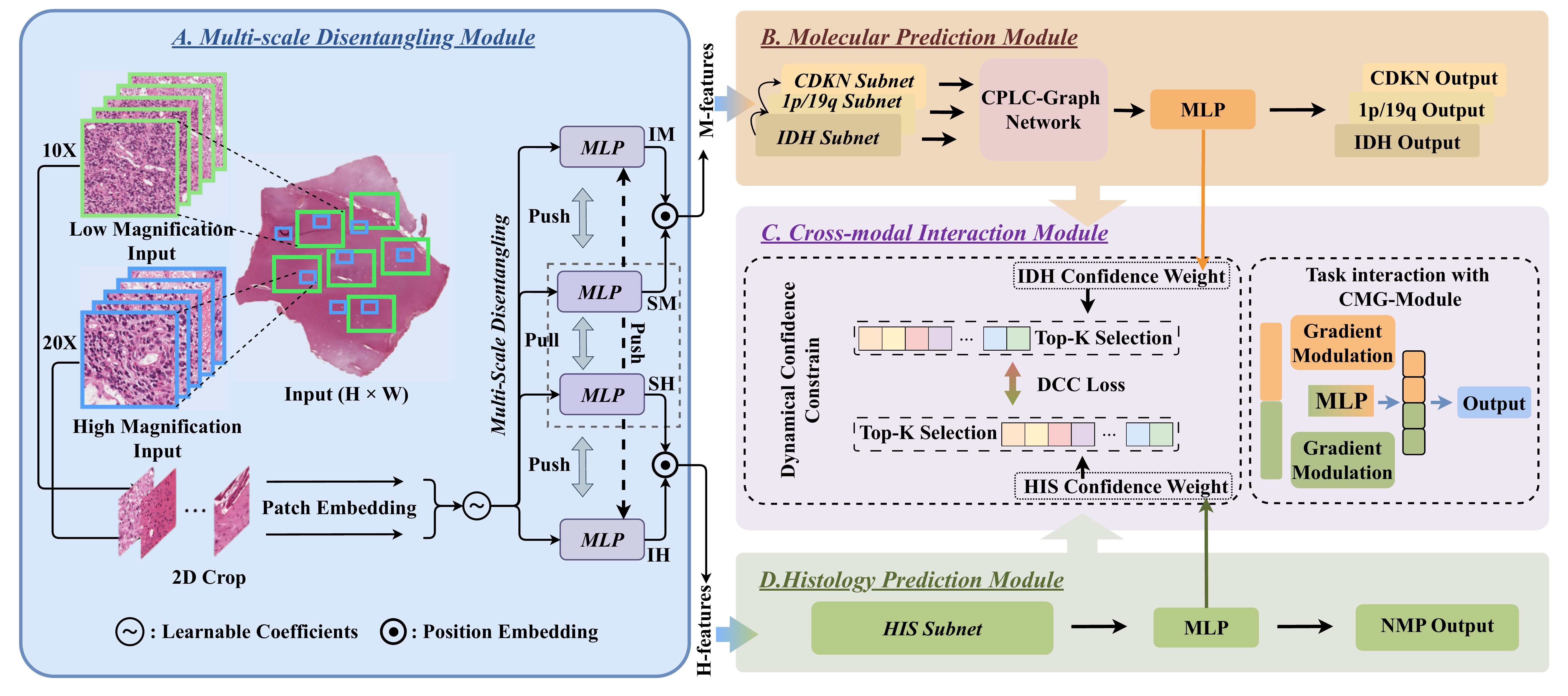}
    \caption{Framework of our M3C2 method, including (A) multi-scale disentangling module, (B) molecular prediction module, (C) cross-module interaction module and (D) histology prediction module. Note that IM, SM, IH and SH denote independent molecular features, shared molecular features, independent histology features and shared histology features, respectively.}
    \label{fig:framework}
\end{figure*}

DL-based tissue-level modelling has been widely researched in cancer diagnosis, promising to enhance the accuracy and efficiency of cancer classification \citep{yang2022pathologic,kanavati2020weakly}. For instance, \cite{lin2018scannet} devised ScanNet, a fast and dense scanning framework utilizing asynchronous sample prefetching and hard negative mining for efficient breast cancer detection. Additionally, \cite{yang2022pathologic} proposed FA-MSCN, a feature aligned multi-scale convolutional network, to achieve liver tumour diagnosis. FA-MSCN integrates features from multiple magnifications to improve the performance by referencing more neighboring information. Similarly, \cite{kanavati2020weakly} designed an end-to-end, high-generalizable, weakly supervised deep convolutional network, which achieved great performance in lung cancer classification. Despite the success in many cancers, DL-based algorithms still face significant challenges in brain tumors, particularly glioma, due to its remarkable heterogeneity, complex diagnostic criteria and limited samples \citep{perry2016histologic,louis20212021}.

\subsection{Automatic glioma grading with WSIs}

In recent years, several models \citep{zhang2022mutual,pei2021deep,lv2024insight} have been proposed to classify glioma based on WSIs. For instance, \cite{zhang2022mutual} designed a mutual contrastive low-rank learning (MCL) scheme, where formalin-fixed paraffin-embedded (FFPE)-based  and frozen-based WSIs are integrated for better glioma grading performance. Additionally, \cite{pei2021deep} proposed a novel DL-based classification method that fuses genomic features with cellularity features for glioma prediction. However, most existing DL methods are based on the obsolete glioma taxonomy criteria \citep{louis20072007,louis20162016}, which only focuses on histology features, such as NMP. Therefore, it is in demand to develop DL-based methods consistent with the latest glioma diagnosis criteria, which jointly concsider histology and molecular markers in the taxonomy pipeline.

Most recently, a few DL-based methods \citep{hollon2023artificial,charm,lv2024insight} have achieved automatic glioma diagnosis using the up-to-date criteria. For example, \cite{hollon2023artificial} proposed a hierarchical vision transformer-based method to separately predict molecular markers and histology features, which are then combined for final diagnosis. Similarly, \cite{lv2024insight} devised a ResNet-50-based DL structure with a novel patch selection strategy for classifying glioma according to the latest criteria. Despite their potentials, significant limitations remain in existing methods: \textbf{(i)} Current methods are incapable of achieving efficient interaction between the histology features and molecular markers for clinically interpretable diagnosis; \textbf{(ii)} The diagnosis scheme used in existing works are incomplete or inconsistent with the latest criteria. For instance, in \citep {hollon2023artificial}, grade 4 astrocytoma, a newly defined class in WHO 2021 criteria, was ignored in the training process. 
\textbf{(iii)} Most models were trained with a single type of WSIs, i.e., either FFPE or frozen tissue sections, whereas in real-world practice, both FFPE- and   frozen-based WSIs are commonly used for diagnosis in different clinical scenarios. To address these gaps, the proposed M3C2 achieves explicit interaction between histology features and molecular markers aligned with the WHO 2021 criteria and trained on both FFPE and frozen sections.

\vspace{-.5em}
\subsection{Integration of histology and molecular markers}

\noindent\textbf{Multi-modal learning for precision oncology.}
Recently, joint modeling of multi-modal data of histology and molecular markers has achieved progress in precision oncology, advancing AI towards practical clinical application in various cancers \citep{zhang2024prototypical, wei2023multi}.
Multi-modal algorithms are generally categorized into three types: early, intermediate, and late fusion. Early fusion integrates raw data or extracted features before inputting them into the proposed algorithms. Intermediate fusion \citep{CMAT, jaume2024modeling} modulates multi-modal feature embeddings at specific fusion modules or different layers within the model. Late fusion \citep{shao2019integrative} involves modality-specific models that aggregate predictions from each modality for the final classification. For example, \cite{CMAT} proposed an intermediate fusion framework (CMAT), which explores intrinsic associations between histopathology images and gene expression profiles through a cross-modal attention module and a modality alignment and fusion module. Similarly,  \cite{jaume2024modeling} developed a late fusion framework named SURVPATH, a memory-efficient multimodal transformer that integrates gene pathways and histopathology images for pan-cancer survival prediction.

For gliomas, several multi-modal data fusion methods have  been proposed \citep{chen2020pathomic, qiu2024dual}. Specifically, \cite{chen2020pathomic} introduced the Pathomic Fusion method that integrates gene expression  and WSI for jointly predicting survival and glioma classification. \cite{qiu2024dual} devised a disentangled multi-modal framework that can handle both complete and incomplete samples for glioma diagnosis and prognosis.  Despite the effectiveness, most current multi-modal methods face practical limitations in clinical scenarios, as they  rely on the input of molecular information, which is expensive and time-consuming to obtain. This dependence hampers their applicability and widespread adoption in clinical practice.

To tackle this challenge, \cite{xing2022discrepancy} proposed a novel discrepancy and gradient-guided modality distillation framework. In this approach, WSIs and gene expressions are combined during training, while the branch for gene expressions is distilled during testing. However, despite this innovative approach, modality distillation-based multi-modal learning methods are fall short in  generating molecular markers. This limitation renders them impractical for clinical use according to the latest diagnosis criteria.

\noindent\textbf{Multi-task learning for precision oncology}
Multi-task learning shows promise to improve on existing methods, which can predict both histology features and molecular markers only using WSIs as input. For instance, \cite{coudray2018classification} proposed an Inception-v3 \citep{szegedy2016rethinking}-based DL method to separately predict both classification and mutation from non–small cell lung cancer histopathology images. However, most existing methods perform multiple tasks separately, without explicit interaction between histology and molecular marker predictions. In comparison, we proposed a novel multi-task learning method with efficient modality interaction via the DCC loss and the specially designed gradient modulation strategy. This approach facilitates explicit task interplay, enhancing the predictive power and clinical applicability of the model by integrating histology and molecular marker more effectively.

\vspace{-.5em}
\subsection{Differences from the conference paper}

This paper is an extended version of our conference paper \cite{deepMO-Glioma} (Nominated for Best Paper Award) with substantial improvements by:
(1) adding a thorough literature review and an extensive discussion.
(2) fulfilling the details of the proposed method and improving the original structure as follows.
We first propose a multi-scale disentangling module to extract multi-magnification WSI features for both histology and molecular marker predictions.
We design a CMG-Modu strategy to harmonize the training process of multiple tasks and achieve cross-modal task interplay.
In addition, an attention mechanism is further involved in our framework to extract more relevant features. 
(3) Consequently, the performance of our method has been improved in all tasks of glioma classification (by 5.6$\%$ in accuracy), molecular markers (e.g., by 4.3$\%$ in accuracy of IDH) and histology (by 4.3$\%$ in accuracy) prediction.
(4) We comprehensively validate the model with diverse experimental settings. First, we perform additional ablation experiments to evaluate the newly proposed modules. Secondly, we add an external validation set of 753 WSIs, verifying the generalizability of our method.  Thirdly, we  evaluate the model performance with varied experimental settings on input magnifications and incorporation of auxiliary tasks. Finally we compare our model with more SOTA methods.

\vspace{-.5em}
\section{Methodology}

\subsection{Framework}

According to the latest glioma diagnosis criteria using both histology and molecular information, it is therefore intuitive to jointly learn the multiple tasks of histology and molecular markers prediction, as well as the final glioma classification, in a unified framework. 
%In this paper, we propose a novel M3C2 method for the primary task of glioma classification, simultaneously handling the multi-modal tasks of histology and molecular markers prediction. 
In this paper, we propose a novel M3C2 method to simultaneously handle these tasks. The framework of M3C2 is shown in Figure \ref{fig:framework}.  
As can be seen, given the cropped multi-scale patches $\{\mathbf{X}^{h}_{i}\}_{i=1}^{N} $ and $\{\mathbf{X}^{l}_{j}\}_{j=1}^{N} \in \mathbb{R}^{N\times H\times W\times 3}$ (with patch number $N$, height $H$, width $W$ and 3 channels of RGB) of 20X and 10X WSI magnification 
as the input, the model can predict 1) molecular markers, including IDH mutation $\hat{l}_{idh} \in \mathbb{R}^{2} $, 1p/19q co-deletion $\hat{l}_{1p/19q} \in \mathbb{R}^{2} $ and CDKN HOMDEL $\hat{l}_{cdkn} \in \mathbb{R}^{2}$, 2) existence of NMP $\hat{l}_{nmp} \in \mathbb{R}^{2} $ and 3) final diagnosis of glioma $\hat{l}_{glio} \in \mathbb{R}^{4}$.
Note that in our 4-class glioma classification task, class 0 to 3 refer to grade 4 GBM, high grade astrocytoma, low grade astrocytoma and oligodendroglioma, respectively.

The structure of M3C2 consists of 4 modules, including multi-scale disentangling module, molecular prediction module, histology prediction module and cross-modal interaction module. Detailed structures are described as follows.

\subsection{Multi-scale disentangling module}
Clinical cancer diagnosis typically relies on  WSI scanning under multiple magnifications (from 20X to 10X), showing morphological features from cellular-level to tissue-level for precision oncology. 
Prior biological knowledge \citep{phan2020high} suggests that the molecular markers and histology features are associated with varied multi-scale features. For instance, the molecular markers could be better reflected by high-magnification features with cellular-level details, while histology features are better captured at the low magnification features at the tissue level, providing information on tissue structures. Hence, we propose a novel multi-scale disentangling module. By disentangling the independent and shared features for molecular and histology tasks, we ensure that the embeddings extracted are more relevant and useful for their respective downstream tasks, rather than relying on a single set of undifferentiated features.
This module can generate representative embeddings for both histology and molecular prediction tasks via the interaction of multi-magnification WSI features.

Given the multi-scale patches $\{\mathbf{X}^{l}_{i}\}_{i=1}^{N}$ and $\{\mathbf{X}^{h}_{j}\}_{j=1}^{N}$ as input, we firstly extract the patch embeddings ${\mathbf{F}^{l} \in \mathbb{R}^{N \times K}}$ and ${\mathbf{F}^{h} \in \mathbb{R}^{N \times K}}$ (with embedding dimension $K$) \citep{mao2023disc} using a pre-trained ResNet-50 \citep{resnet}. The multi-scale extracted embeddings  are then fed into the multi-scale disentangling module, a multi-layer perceptron (MLP)-based network for producing molecular- and histology-oriented features, as shown in Figure \ref{fig:framework}. Specifically, the embeddings $\mathbf{F}^{l}$ and $\mathbf{F}^{h}$ are first weighted  with learnable  coefficients and then concatenated together and finally processed with four separate MLPs to generate both shared and independent features for the molecular ($\mathbf{S}_{\rm m}$ and $\mathbf{I}_{\rm m}$ for shared and independent features) and histology ($\mathbf{S}_{\rm h}$ and $\mathbf{I}_{\rm h}$) prediction tasks. Moreover, we further proposed a cross-modal disentanglement loss to  minimize the disparity among shared representations while maximize that among independent representations. Formally, the disentanglement loss is defined as follows.

\begin{gather}
\label{loss_disen}
\mathcal{L}_{\rm disent}= \frac{\|\mathbf{S}_{\rm m} - \mathbf{S}_{\rm h}\|_2 }{\|\mathbf{I}_{\rm m} - \mathbf{I}_{\rm h}\|_2 + \|\mathbf{I}_{\rm m} - \mathbf{U}_{\rm m}\|_2 + \|\mathbf{I}_{\rm h} - \mathbf{U}_{\rm h}\|_2},
\end{gather}

%we disentangle the module into four components: Private Molecular (PM\mathbf{PM}), Common Molecular (CM\mathbf{CM}), Common Histopathology (CH\mathbf{CH}), and Private Histopathology (PH\mathbf{PH}). 
% molecular features (M-features) Fmole\mathbf{F}_{\rm mole} and histology features (H-features) Fhis\mathbf{F}_{\rm his} 
%The molecular features (M-features) Fmole\mathbf{F}_{\rm mole} are derived from the concatenation of PM(Fl,Fh)\mathbf{PM}(\mathbf{F}^{l}, \mathbf{F}^{h}) and CM(Fl,Fh)\mathbf{CM}(\mathbf{F}^{l}, \mathbf{F}^{h}), while histology features (H-features) Fhis\mathbf{F}_{\rm his} is obtained by concatenating PH(Fl,Fh)\mathbf{PH}(\mathbf{F}^{l}, \mathbf{F}^{h}) and CH(Fl,Fh)\mathbf{CH}(\mathbf{F}^{l}, \mathbf{F}^{h}). These two task-oriented features will be utilized in the following subnet features.}

%To further enhance the model's representation learning capability in distinguished spaces with common and private features, we introduce a constraint Lscale\mathcal{L}_{scale}. This constraint is designed to simultaneously pull the common and private features within the same space while pushing the private features into diverse spaces. The constraint is formulated as:
%待补充
%Lscale\mathcal{L}_{scale}
%The proposed method enhances the comprehensive analysis of histopathological data, potentially improving the accuracy and reliability of brain tumour classification.

\subsection{Attention-based hierarchical multi-task multi-instance learning}

\begin{figure}[t]
    \centering
    \includegraphics[width=1\linewidth]{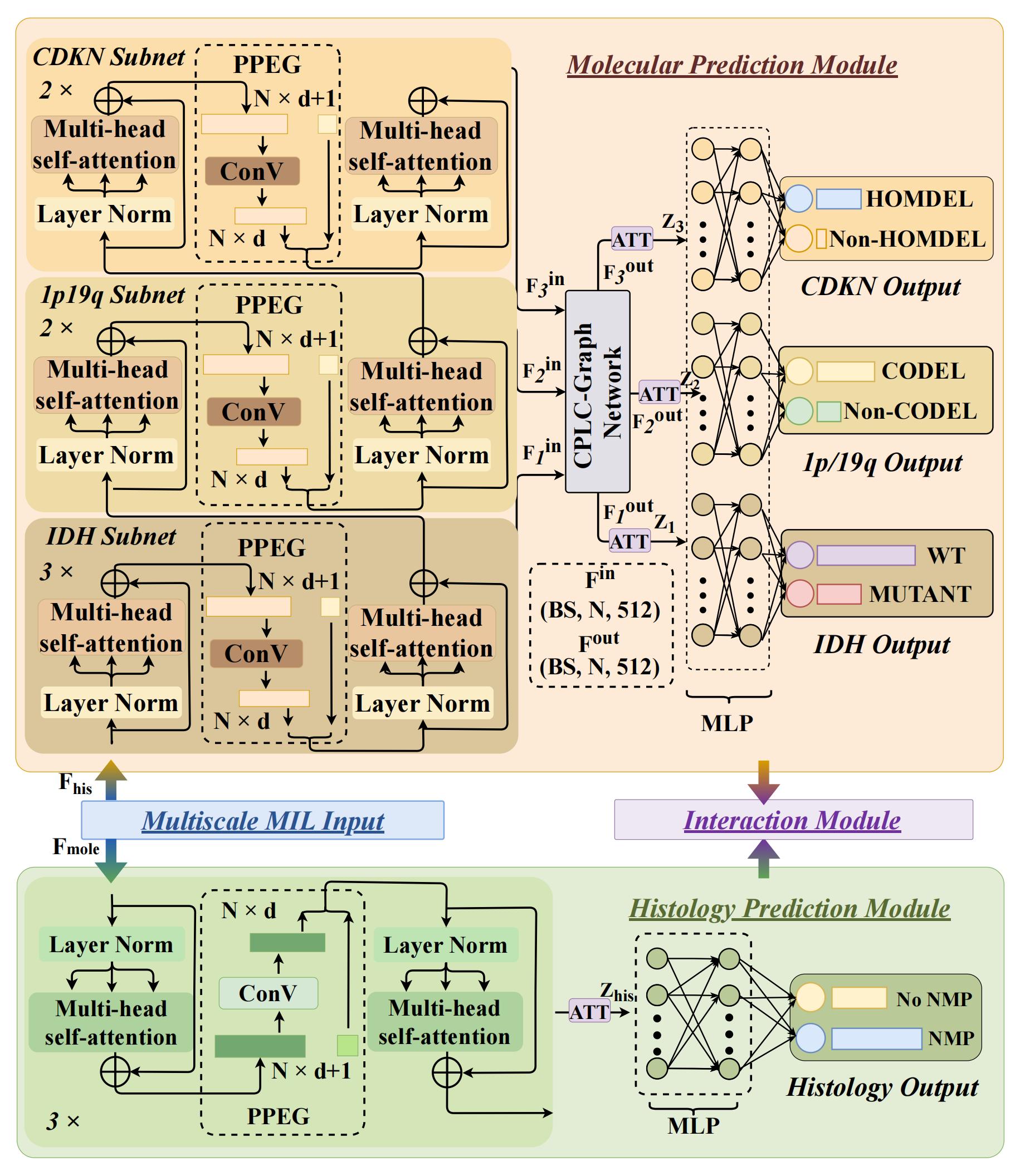}
    \caption{Detailed structure of the proposed molecular prediction module (above) and the histology prediction module (below).}
    \label{fig:hisandmole}
\end{figure}

To further extract molecular and histology information for the multi-modal prediction tasks, we propose an attention-based hierarchical multi-task multi-instance learning (AHMT-MIL) framework. Different from methods using one \citep{zhang2022mutual} or several \citep{xing2022discrepancy} representative patches per slide, AHMT-MIL framework can extract information from N=2,500 patches per WSI on both magnifications via utilizing the MIL  learning paradigm. Of note, for WSIs with patch number$<$ N and $>$ N, we adopt the biological repeat  and 2D average sampling strategy for dimension alignment, respectively. 
Specifically, developed on the top of vision transformer \citep{dosovitskiy2020image}, the AHMT-MIL framework consists of two  modules for
the  histology and molecular prediction tasks. Figure \ref{fig:hisandmole} illustrates the structure of these two modules, which are further detailed as follows.

\noindent\textbf{Histology prediction module.}
As illustrated in Figure \ref{fig:hisandmole}, the histology prediction module takes the H-features $\mathbf{F}_{\rm his}$ as input and outputs the prediction of the histology features of NMP. Specifically,  $\mathbf{F}_{\rm his}$ is firstly processed through three cascaded transformer blocks to extract NMP features, which is then refined with the attention mechanism \citep{ilse2018attention} for more distinctive histology features. Details of the attention mechanism are described below.

 Given a bag of $N$ embeddings $\mathbf{F}=\{\mathbf{f}_{1}, \ldots , \mathbf{f}_{N}\} \in \mathbb{R}^{N \times K}$, the refined embeddings $\mathbf{z} \in \mathbb{R}^{1 \times K}$ can be generated via an MIL pooling as follows:
\begin{equation}\label{eq:weighted_sum}
\mathbf{z} = \sum_{n=1}^{N} a_n \mathbf{f}_{n},
where \;
a_{n} = \dfrac{ \exp\{\mb{w}^{\top} \tanh \big{(} \mb{V} \mathbf{f}_{n}^{\top} \big{)}\}}{{\displaystyle \sum_{j=1}^N} \exp \{ \mb{w}^{\top} \tanh \big{(}\mb{V} \mathbf{f}_{j}^{\top}\big{)} \} }.
\end{equation}
In equation \eqref{eq:weighted_sum}, $\mb{w} \in \mathbb{R}^{L \times 1}$ and $\mb{V} \in \mathbb{R}^{L \times K}$  are learnable parameters, while $\mathrm{tanh(\cdot )}$ denotes element-wise hyperbolic tangent function. Finally, the refined embeddings $\mathbf{z}$ is fed to the MLP for the histology prediction task.

\begin{figure}[t]
    \centering
    \includegraphics[width=1\linewidth]{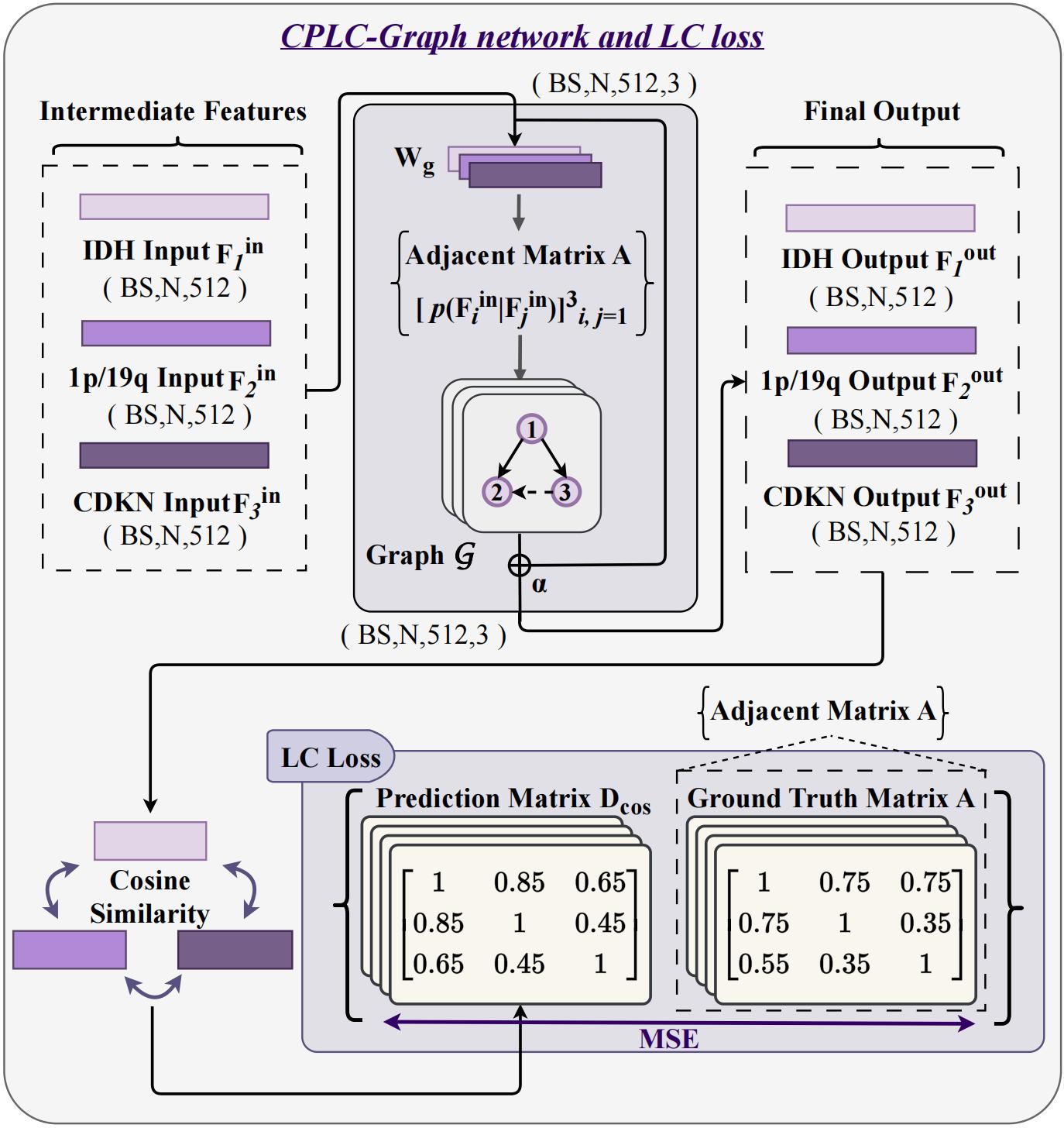}
    \caption{Detailed structure of the proposed CPLC-Graph network and the LC loss.}
    \label{fig:cplc}
\end{figure}

\noindent\textbf{Molecular prediction module.} 
Taking M-features $\mathbf{F}_{\rm mole}$ from the multi-scale disentangling module as input, the molecular prediction module is developed for the multi-label classification \citep{liu2021emerging} of  the three molecular markers. As shown in Figure \ref{fig:hisandmole}, the molecular prediction module consists of three sequential subnets, extracting features for IDH, 1p/19q and CDKN markers, respectively.  Specifically, each subnet is composed of several cascaded transformer blocks, with the block number of 3, 2 and 2 for IDH, 1p/19q and CDKN, respectively\footnote{ Of note, IDH is a relative upstream genetic event, so that it is in the forefront of the feature extracting sequence.}. In addition, similar to histology prediction, attention mechanism is also utilized in the MIL process of for the molecular prediction task, allowing for the extraction of representative patches for these three markers. Finally, we further devised a graph-based network to model the intrinsic correlation of the three molecular markers, which is introduced as follows.

\subsection{Co-occurrence probability-based, label-correlation graph}

In classifying molecular markers, including IDH, 1p/19q and CDKN,  current MLC methods based on label correlation might overlook the co-occurrence of these labels. To address this, we propose a co-occurrence probability-based label-correlation graph (CPLC-Graph) network and a label correlation (LC) loss, which are designed for intra-omic modeling of the co-occurrence probabilities among the three markers.

\noindent\textbf{1) CPLC-Graph network:} As shown in Figure \ref{fig:cplc}, CPLC-Graph is defined as $ \mathcal{G}=(\mathbf{V},\mathbf{E})$, where  $\mathbf{V}$  indicates the nodes and $\mathbf{E}$ represents the edges. 
 Given the intermediate features in predicting the three molecular markers  $\mathbf{F}^{\rm in}=[\mathbf{F}^{\rm in}_{i}]_{i=1}^{3} \in \mathbb{R}^{3\times N\times K}$ as input nodes, we then construct a co-occurrence probability based correlation matrix $\mathbf{A} \in \mathbb{R}^{3\times 3}$ to capture the relationships among these node features. Note that $\mathbf{A}$ is a shared value calculated using our internal large-scale TCGA dataset.
Additionally, a weight matrix $\mathbf{W}_{g} \in \mathbb{R}^{K\times K}$ is utilized to update $\mathbf{F}^{\rm in}$. The output nodes $\mathbf{F}^{\rm mid} \in \mathbb{R}^{3\times N\times K}$ are then computed using a single layer of a graph convolutional network, formulated as follows.

\begin{gather}
\label{E1}
\mathbf{F}^{\rm mid} = \mathcal{\delta} (\mathbf{A} \mathbf{F}^{\rm in} \mathbf{W}_{g}), \\
\text{where} \mathbf{A} = [A_{i}^{j}]_{i,j=1}^{3}, A_{i}^{j} = \frac{1}{2}\big (p (\mathbf{F}^{\rm in}_{i}|\mathbf{F}^{\rm in}_{j})+p (\mathbf{F}^{\rm in}_{j}|\mathbf{F}^{\rm in}_{i})\big).
\end{gather}

In the above equation, $\mathcal{\delta} (\cdot)$ represents  an activation function and $p (\mathbf{F}^{\rm in}_{i}|\mathbf{F}^{\rm in}_{j})$ indicates  the probability of the status of $i$-th marker given the status of $j$-th marker. Additionally, residual structure is employed  to produce  the final output $\mathbf{F}^{\rm out}$ of CPLC-Graph network, which is defined as  
$\mathbf{F}^{\rm out} = \alpha \mathbf{F}^{\rm mid} + (1-\alpha) \mathbf{F}^{\rm in}, $
where $\alpha$ is a graph balancing hyper-parameter.

\noindent\textbf{2) LC loss:} To fully leverage the co-occurrence probability of different molecular markers, we introduce the LC loss, which constrains the similarity between any two output molecular markers, $\mathbf{F}^{\rm out}_{i}$ and $\mathbf{F}^{\rm out}_{j}$, to align with their corresponding co-occurrence probability $A_{i}^{j}$. This approach aims to better capture the intrinsic associations among various genomic alterations. Formally, the LC loss is defined as follows:

\begin{gather}
\label{loss0}
\mathcal{L}_{\rm LC}= \mathcal{MSE} (\mathbf{A},\mathbf{D}_{\rm cos}), \\
\text{where}\; \mathbf{D}_{\rm cos} =[D_{cos}^{i,j}]_{i,j=1}^{3}, D_{cos}^{i,j}= \frac{(\mathbf{F}^{\rm out}_{i})^\top \mathbf{F}^{\rm out}_{j}} { \left\| \mathbf{F}^{\rm out}_{i} \right\| \left\|\mathbf{F}^{\rm out}_{j} \right\| }. 
\end{gather}

In \eqref{loss0}, $\mathcal{MSE}$ is the function of mean square error, while $D_{cos}^{i,j}$ denotes the cosine similarity of features $\mathbf{F}^{\rm out}_{i}$ and $\mathbf{F}^{\rm out}_{j}$.

\subsection{Cross-modal interaction module}

To achieve efficient interaction between histology and molecular markers prediction tasks, we further proposed the cross-modal interaction module  with the specially designed CMG-Modu training strategy and dynamic confidence constrain loss, which are detailed as follows.

\subsubsection{CMG-Modu learning strategy}

\begin{figure}[!t]
    \centering
    \includegraphics[width=1\linewidth]{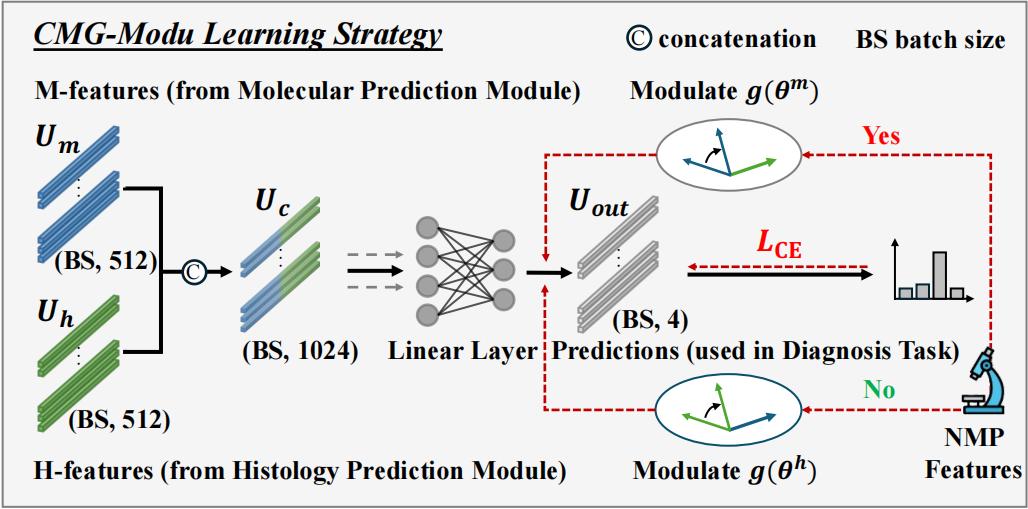}
    \caption{Illustration of the CMG-Modu learning strategy}
    \label{fig:CGM}
\end{figure}

%Due to the updating of diagnostic criteria for brain tumours, molecular-pathology diagnosis has shown success in glioma diagnosis. 
% The penetration of molecular information calibrates the histopathology-based diagnostic results, which promotes precision neuro-oncology. For example, researchers find that prognostic result may be unexpected under the 2016 WHO criteria for low-grade glioma, while more accurate outcomes can be achieved with the introduction of molecular markers, such as IDH mutation and 1p/19q co-deletion. 
According WHO 2021 criteria, the molecular markers and histology features are closely integrated for the glioma diagnosis. For instance, patients previously diagnosed as high grade based on \textit{histology features of NMP} are consistently considered as high grade in the updated criteria. In contrast, the grade of patients without NMP features turns to be determined by \textit{molecular markers} in the latest diagnostic scheme \citep{berger2022world}.
Inspired by this clinical prior, we propose the cross-modal gradient modulation (CMG-Modu) learning strategy, as shown in Fig. \ref{fig:CGM}, to facilitate the interplay between the molecular markers and histology prediction tasks. Generally, the optimization process is dynamically calibrated by the gradient of the dominant task, i.e., either predicting molecular markers or histology features of NMP. Detailed design of the CMG-Modu is as follows.

As illustrated in Fig. \ref{fig:CGM}, we firstly concatenate  the extracted histology features $\mathbf{U}_h$ and molecular features $\mathbf{U}_m$ to generate the multi-modal features $\mathbf{U}_c$, which is further fed to a fully connected layer for the glioma classification.
During the training process, the gradients in optimizing the multi-modal prediction tasks are modulated by the dominant task in terms of the labels of NMP, thus calibrating the prediction bias. Specifically, we modulate the gradients of molecular markers prediction task ($g(\theta^h)$) using those of histology prediction task when the histology label within a minibatch\footnote{Note that the histology label within a minibatch is obtained by majority voting.)} is NMP positive, and vice versa. Formally, the modulated gradients for each task is defined as follows:
\begin{equation} \label{e:gm1}
    \begin{cases}
        \ \tilde{g}(\theta^h) \ = \ \rho(\Phi (g(\theta^h), \ g(\theta^m))), & NMP \ negative, \\
        \ \tilde{g}(\theta^m) \ = \ \rho(\Phi (g(\theta^m), \ g(\theta^h))), & NMP \ positive,
    \end{cases}
\end{equation}
where $\rho(\mathbf{\cdot})$ is a scaling function to fix gradient scale (\cite{wang2024mgiml}), while $\Phi(\mathbf{a},\mathbf{b})$ represents the projection of vector $\mathbf{a}$ onto the direction perpendicular to vector $\mathbf{b}$, which can be calculated as:
\begin{equation}
\text{Proj}_{\perp b} \mathbf{a} = \mathbf{a} - \text{Proj}_{b} \mathbf{a}  
\end{equation}
where \(\text{Proj}_{b} \mathbf{a}\) is the projection of vector $\mathbf{a}$ onto vector $\mathbf{b}$:
\begin{equation}
\text{Proj}_{b} \mathbf{a} = \frac{\mathbf{a} \cdot \mathbf{b}}{\|\mathbf{b}\|^2} \mathbf{b}
\end{equation}

Moreover, the detailed model learning process with CGM-Modu strategy can be found in Algorithm \ref{al:gm}. As can be seen, the cross-modal optimization reflecting the up-to-date diagnostic criteria can relieve the modality-specific bias and enhances the final glioma classification performance, as validated in Section \ref{section_ablation}.

\begin{algorithm}[t] 
    \caption{Training process with CMG-Modu  strategy} \label{al:gm}
    \begin{algorithmic}[1]
        \REQUIRE{Paired histopathology features and molecular markers features $\{\mathbf{U}_h, \mathbf{U}_m \}$ for up-to-date brain tumor diagnosis; Model parameters $\Theta = \{\theta^h, \theta^m\}$; Multi-modal classifier $\mathcal{D}$; Task label $Y$; Histopathology-based grading label $T$.}
        
        \ENSURE{The optimal model parameters $\Theta^*$}
        
        %\textbf{The CMG-Modu learning strategy}
        \WHILE{$\Theta$ doesn't reach convergence} 
        \FOR{each $\mathbf{u}_h$ and $\mathbf{u}_m$ $\in$ $\{\mathbf{U}_h, \mathbf{U}_m \}$}
        \STATE Concatenate $\mathbf{u}_h$ and $\mathbf{u}_m$ to get features $\mathbf{u}_c$
        \STATE Minimize ${\mathcal{L}_{\rm{CE}}}(Y, \mathcal{D}(\mathbf{u}_c ; \theta^h, \theta^m)$

        \IF {T belongs to high grade}
        \STATE Calibrate molecular gradients using Eq. \eqref{e:gm1}
        \ELSE
        \STATE Calibrate histopathology gradients using Eq. \eqref{e:gm1}
        \ENDIF
        
        \ENDFOR
        \ENDWHILE
        \STATE The $\mathcal{D}$ is well-trained by CMG-Modu.

    \end{algorithmic}
\end{algorithm}

\begin{table*}[!t]%
\scriptsize
    \centering
     \caption{Mean  values in terms of percentage for glioma classification metrics by our and other methods over the internal and external validation datasets.
     }
     \label{Performance1}
    \begin{tabular}{|c|ccc|ccccc|ccccc|} \hline 
         Approaches&  \multicolumn{3}{|c|}{Attributes}&  \multicolumn{5}{|c|}{Evaluation on internal dataset}&  \multicolumn{5}{|c|}{Evaluation on external dataset}\\ \hline 
         &  MTL*&  CMM**&  For glioma&  Acc.&  Sen.&  Spec.&  AUC&  $\mathrm{F_{1}}$-score&  Acc.& Sen.& Spe.& AUC&$\mathrm{F_{1}}$-score\\ \hline 
         VGG&  &  &  &  71.1&  71.1&  85.2&  84.9&  71.1&  89.4& 89.4& 81.6& 91.8&89.4\\
 Inception& & & & 60.2& 60.2& 84.3& 81.4& 60.2& 67.2& 67.2& 99.5& 84.5&67.2\\ 
         DenseNet&  &  &  &  63.2&  63.2&  \textbf{91.6}&  84.8&  63.2&  68.3& 68.3& 99.3& 73.7&68.3\\ 
         AlexNet&  &  &  &  53.3&  53.3&  89.0&  71.0&  53.3&  80.2& 80.2& 99.6& 89.9&80.2\\
 ResNet& & & & 65.5& 65.5& 52.7& 80.4& 65.5& 55.3& 55.3& 63.5& 94.6&55.3\\ 
         \hline 
         ABMIL& & & \checkmark& 69.4& 69.4& 82.3& 85.9& 69.4& 93.1& 93.1& 81.7& 97.9&93.1\\
         CLAM&  &  &  \checkmark&  63.5&  63.5&  91.1&  83.7&  63.5&  82.2& 82.2& \textbf{99.9}& 86.4&82.2\\ 
         TransMIL&  &  &  \checkmark&  66.1&  66.1&  84.0&  85.6&  66.1&  82.3& 82.3& 69.3& 93.1&82.3\\
         \hline 
         Charm&  &  &  \checkmark&  57.6&  57.6&  83.0&  79.6&  57.6&  54.3& 54.3& 81.0& 64.4&54.3\\ 
 Deepglioma& & & \checkmark& 55.3& 55.3& 84.9& 79.9& 55.3& 66.9& 66.9& 99.7& 84.6&66.9\\
         \hline 
 MCAT& & \checkmark& \checkmark& 68.0& 68.0& 89.3& 87.7& 68.0& -& -& -& -& -\\
 CMTA& & \checkmark& \checkmark& 66.3& 66.3& 88.8& 87.0& 66.3& -& -& -& -& -\\ 
         Wang \textit{et al.}  $\dag$&  \checkmark&  \checkmark&  \checkmark&  73.0 &  73.0 &  82.3 &  87.2 &  73.0 
&  95.7& 95.7& 51.2& 99.7 &95.7\\ \hline 
         \textbf{Ours}&   \checkmark&   \checkmark&   \checkmark&  \textbf{78.6}&  \textbf{78.6}&  85.4&  \textbf{88.7}&  \textbf{78.6}&  \textbf{98.1}& \textbf{98.1}& 69.5& \textbf{99.7}&\textbf{98.1}\\ \hline
 \multicolumn{14}{l}{* MTL refers to multi-task learning. ** CMM denotes cross-modal modelling. $\dag$  Wang \textit{et al.} is the conference version of our method.}\\
    \end{tabular}
\end{table*}

\subsubsection{Dynamic confidence constrain}
Additionally, we have further developed a dynamic confidence constrain (DCC) strategy to model the interaction between molecular markers and histological features. The DCC incorporates known clinical co-occurrence relationships between molecular and histology features, such as IDH wildtype and NMP. IDH wildtype is used as a major diagnostic marker for glioblastoma in the current clinical paradigm, while historically, the diagnosis of glioblastoma is purely based on NMP. In addition, studies have shown that over 95$\%$ of patients with NMP are IDH wildtype \citep{alzial2022wild}, highlighting the strong co-occurrence relationship between the two markers. This relationship underpins the potential performance benefits of leveraging this co-occurrence for molecular prediction. This design not only enhances the interpretability of molecular predictions (Figure \ref{fig:visualization}) but also improves prediction performance.
Specifically, inspired by \citep{zhang2022dtfd}, the confidence weights for IDH wildtype $\mathbf{C}_{\rm wt} = [c_{\rm wt}^{n}]_{n=1}^{N} \in \mathbb{R}^{N \times 1}$ can be defined as:

\begin{gather}
 \label{dcc}
 \mathbf{C}_{\rm wt} = (\mathbf{F}^{\rm out}_{1} \cdot \mathbf{\hat{z}}_{1}) \otimes \mathbf{w}_{\rm wt}, \text{where} \; \mathbf{\hat{z}}_{1} \in \mathbb{R}^{N \times K} \xleftarrow{\text{repeat}} \mathbf{z}_{1} \in \mathbb{R}^{1 \times K}
 \end{gather}

In \eqref{dcc}, $\cdot$ denotes the dot product, $\otimes$ represents Hadamard product, $\mathbf{z}_{1}$ is refined IDH embeddings with attention mechanism and  $\mathbf{w}_{\rm wt}$ is the weights in the MLP for IDH wildtype.
We then reorder $[c_{\rm wt}^{n}]_{n=1}^{N}$ to $[\hat{c}_{\rm wt}^{n}]_{n=1}^{N}$ based on their values. Similarly,  we obtain $\mathbf{C}_{\rm nmp} = [\hat{c}_{\rm nmp}^{n}]_{n=1}^{N}$ for NMP confidence weights.

Using ordered confidence weights, we constrain the prediction networks of histology and molecular markers to concentrate  on the WSI areas important for both predictions, thereby modeling cross-modal interactions. Specifically, the confidence constraint is achieved by a novel DCC loss, which focuses on the top $M$ important patches for both predictions. Formally, the DCC loss in $p$-th training epoch is defined as:
\begin{equation}
\label{loss1}
\mathcal{L}_{\rm DCC}= \frac{1}{2M_p} \sum\nolimits_{m=1}^{M_p} \big( \mathcal{S}(\hat{c}_{wt}^{m},\mathbf{C}_{\rm nmp})+\mathcal{S}(\hat{c}_{nmp}^{m},\mathbf{C}_{\rm wt})\big),
\end{equation} 
where $\mathcal{S}(\hat{c}_{wt}^{m},\mathbf{C}_{\rm nmp})$ is the indicator function that takes the value 1 when the $k$-th important patches of IDH widetype is among the top $M_p$ important patches for NMP, and vice versa. In addition, to enhance the learning process with DCC loss, we employ  a curriculum-learning-based training strategy dynamically focusing on hard-to-learn patches,  identified as those  with higher decision importance weight, as patches with lower confidence weight, such as those with fewer nuclei, are typically easier to learn for  both tasks. Hence, $M_p$ is further defined as

\begin{equation}
\label{eq2}
M_p=M_0 \beta^{\lfloor\frac{p}{p_0} \rfloor }.
\end{equation} 
In \eqref{eq2}, $M_0$ and $p_0$ are hyper-parameters to adjust  $\mathcal{L}_{\rm DCC}$ in training process.
\section{Experiment}

\begin{figure*}
    \centering
    \includegraphics[width=.95\linewidth]{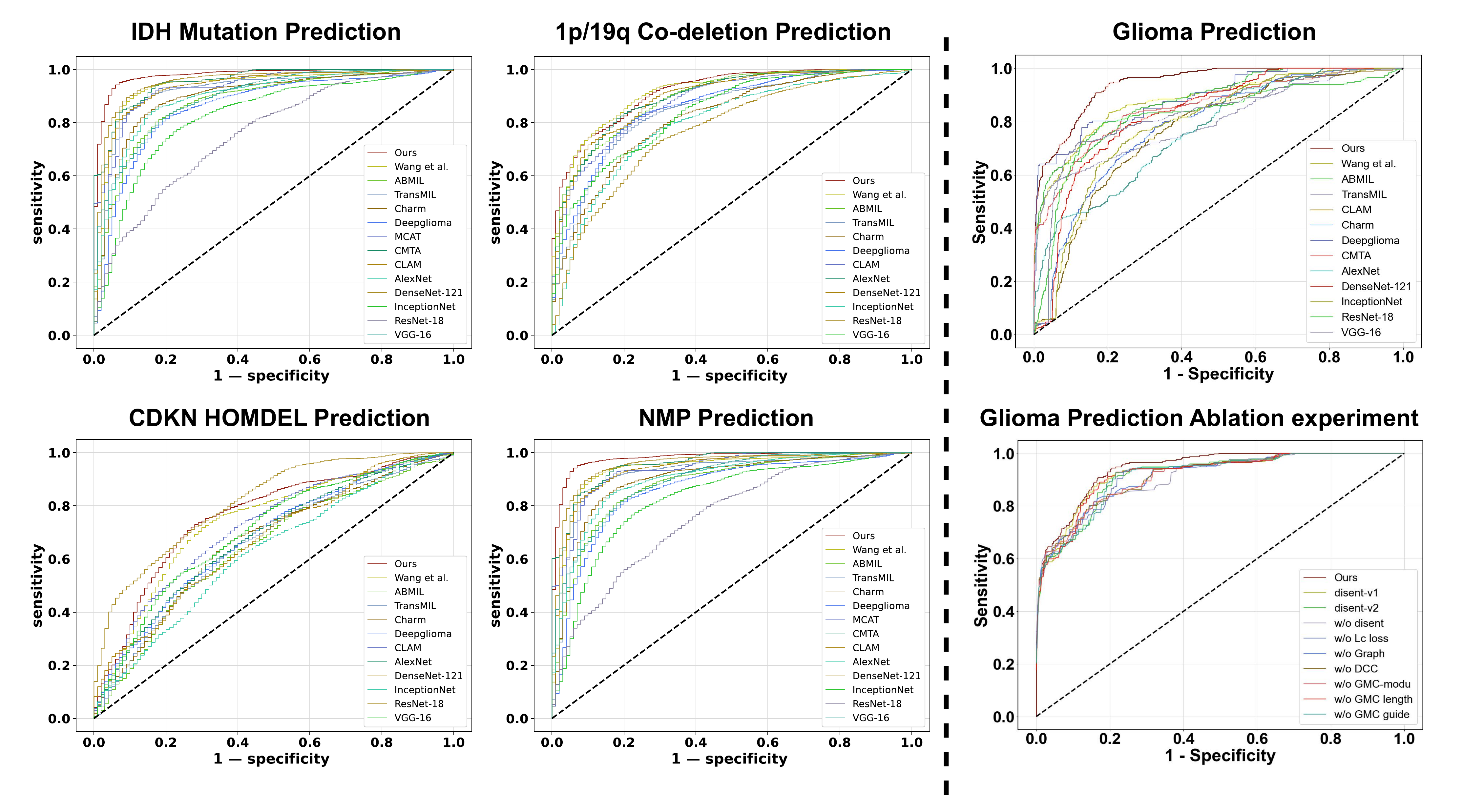}
    \caption{ROCs of our model, comparison and ablation models for predicting IDH,
1p/19q, CDKN, NMP and Glioma.}
    \label{fig:roc}
\end{figure*}

\subsection{Datasets and training labels}

\noindent\textbf{Dataset preparation.} 
Overall, three public datasets, i.e., TCGA GBM-LGG \citep{tomczak2015review}, CPTAC \citep{verdugo2022update} and IvYGAP \citep{eberhart2020spatial}, are involved in our study. In all datasets, we remove the WSIs of low quality or lack of labels, consistent with settings in \citep{lu2021data}. Totally, we include 3,578 WSIs from 1,054 cases. Following \citep{nguyen2023combining}, the two multi-regional TCGA GBM-LGG and CPTAC datasets are further merged as the meta dataset, i.e., internal dataset, for better learning performance. Additionally, IvYGAP  dataset is used as an external validation dataset. See more details of dataset preparation in section 1 of supplementary material.

\noindent\textbf{Training labels.} Original labels for molecular markers and histology of WSIs are obtained from the three used datasets. According to the up-to-date WHO criteria \cite{louis20212021},  we generate the classification labels for each case as grade 4 glioblastoma (defined as IDH wildtype), oligodendroglioma (defined as IDH mutant and 1p/19q co-deletion), grade 4 astrocytoma (defined as IDH mutant, 1p/19q non co-deletion with CDKN HOMDEL or NMP), or low-grade astrocytoma (other cases).
Detailed diagnosis pipeline of glioma based on WHO 2021 can be found in Figure 1 in supplementary material.

\subsection{Implementation details}
The proposed M3C2 model is trained on the internal training set for 50 epochs, with a batch size of 6 and a learning rate of 0.003 using the Adam optimizer \citep{kingma2014adam}, alongside weight decay. Key hyperparameters are detailed in Table 1 of the supplementary material. All hyperparameters are fine-tuned to achieve optimal performance on the internal validation set. Experiments are conducted on a computer equipped with an AMD EPYC 7H12 CPU @1.70GHz, 1TB RAM, and 8 Nvidia Tesla V100 GPUs. Our method is implemented in PyTorch within a Python environment. 

\subsection{Performance evaluation}

\noindent\textbf{Glioma classification.} 
We compare our method with thirteen  other SOTA methods, including five commonly-used image classification methods (AlexNet \citep{alexnet}, ResNet-18 \citep{resnet}, InceptionNet \citep{Inception}, MnasNet\citep{mnasnet}, DenseNet-121 \citep{densenet}), three SOTA MIL framework (ABMIL \citep{ilse2018attention}, TransMIL \citep{transmil} and CLAM \citep{lu2021data}), two SOTA multi-modal learning methods (MCAT \citep{MCAT}, CMTA \citep{CMAT}) three SOTA methods specially designed for glioma classification based on WHO 2021 (Charm \citep{charm}, Deepglioma \citep{khalighi2024artificial} and DeepMO-Glioma \citep{deepMO-Glioma}). Note that DeepMO-Glioma is the conference version of our method.
All the compared methods are trained by the classification labels of glioma, which are defined by integrating molecular markers with histology, thereby demonstrating the fairness of our comparison.
In our experiments, we apply five metrics\footnote{Note that we use micro-average for all metrics in glioma classification.} to assess glioma classification: accuracy, sensitivity, specificity, AUC and $\mathrm{F_{1}}$-score. 
The middle panel of Table \ref{Performance1} shows that shows that M3C2 performs the best on the internal dataset, achieving at least 5.6\%, 5.6\%, 1.5\% and 5.6\%
improvement over other models in accuracy, sensitivity, AUC and $\mathrm{F_{1}}$-score, respectively.
These significant improvements are mainly due to the following aspects. 
 (1) We design a multi-scale disentangling module to extract multi-magnification WSI features for both the molecular marker and histology prediction tasks
 (2) We propose a CPLC-Graph network for efficient intra-omic modelling of the intrinsic correlation of the molecular markers.
 (3) We devise the DCC loss and CMG-Modu training strategy for the cross-modal interaction of histology and molecular markers.
 The corresponding analysis can be found in the ablation study
of Section \ref{section_ablation}.
Additionally, Figure \ref{fig:roc} plots the ROCs of all models, demonstrating the superior performance of our model over other comparison models.

\begin{table*}[t]
\scriptsize
    \centering
    \caption{Mean  values in terms of percentage for predicting molecular markers and histology over the internal dataset.}
    \label{Performance2}
    \begin{tabular}{|c|c|clcccccccccccc|} \hline 
         Task&  Metrics&  Ours & Wang \textit{et al.} &  ABMIL&  TransMIL&   CLAM& Charm&  Deepglioma& MCAT& CMTA&Alex.&Dense.&   Inception&  Res.& VGG\\ \hline 
         \multirow{5}{*}{\rotatebox{90}{IDH}}&  Acc.&  \textbf{88.8}&84.5 
&  80.9&  84.2&   84.2& 82.6&  80.9& -& -&83.6&74.0&  75.7&  68.4& 83.2\\  
         &  Sen.&  82.6  &78.5 
&  76.0&  84.3&   78.5& 82.6&  76.9& -& -&79.3&\textbf{92.6}&  79.3&  39.7& 81.0\\  
         &  Spec.&  \textbf{92.9}&88.5 
&  84.2&  84.2&   88.0& 82.5&  83.6& -& -&86.3&61.7&  73.2&  87.4& 84.7\\ 
         &  AUC&  \textbf{95.0}&92.1 
&  90.5&  89.7&   92.7& 87.8&  87.3& -& -&91.9&86.5&  78.9&  82.0& 88.9\\ 
         &  $\mathrm{F_{1}}$&  \textbf{85.5}&80.2 
&  76.0&  81.0&   79.8& 79.1&  76.2& -& -&79.3&73.9&  72.2&  50.0& 79.4\\ \hline 
 \multirow{5}{*}{\rotatebox{90}{1p19q}}& Acc.
& 80.6  &\textbf{82.2}& 79.0& 78.3&  78.6& 73.7&  79.9& -& -&81.6&78.6& 74.3& 72.0&72.0\\ 
 & Sen.
& \textbf{81.7}&63.4 
& 46.8& 40.2&  63.4& 76.8&  72.0& -& -&52.4&57.3& 62.2& 48.3&84.1\\ 
 & Spec.
& 80.2  &89.2 
& 87.1& 92.3&  84.2& 72.5&  82.9& -& -&\textbf{92.3}&86.5& 78.8& 78.9&67.6\\ 
 & AUC
& \textbf{90.9}&89.1 
& 81.6& 76.4&  85.1& 82.0&  83.7& -& -&86.8&85.4& 76.6& 69.0&81.5\\  
 & $\mathrm{F_{1}}$& \textbf{69.4}&65.8 
& 47.3& 50.0&  61.5& 61.2&  65.9& -& -&60.6&59.1& 56.7& 43.9&61.9\\ \hline
 \multirow{5}{*}{\rotatebox{90}{CDKN}}& Acc.
& \textbf{72.7}&71.1& 63.5& 63.8&  65.1& 64.1&  62.2& -& -&62.8&61.5& 60.5& 71.6&64.8\\ 
 & Sen.
& 76.3  &79.4 
& 59.0& 68.8&  \textbf{81.9}& 58.1&  46.3& -& -&75.6&50.6& 41.3& 48.3&58.8\\  
 & Spec.
& 68.8  &61.8 
& 69.7& 58.3&  46.5& 70.8&  79.9& -& -&48.6&73.6& \textbf{81.9}& 78.4&71.5\\  
 & AUC
& \textbf{76.0}&75.0 
& 66.2& 65.7&  73.6& 67.8&  70.7& -& -&66.8&71.8& 67.5& 72.8&70.7\\ 
 & $\mathrm{F_{1}}$& \textbf{74.6}&74.3 
& 65.0& 66.7&  71.2& 63.1&  56.3& -& -&68.2&58.1& 52.4& 43.6&63.7\\ \hline 
 \multirow{5}{*}{\rotatebox{90}{NMP}}& Acc.
& \textbf{94.4}&90.1 
& 82.4& 87.5&  86.2& 84.5&  80.9& 86.0& 86.6&85.2&89.8& 78.3& 69.1&81.6\\ 
 & Sen.
& 93.4 
&88.0 
& \textbf{93.6}& 87.4&  82.0& 82.6&  82.6& 75.9& 77.8&80.2&90.4& 73.7& 68.9&71.9\\ 
         &  Spec.
&  \textbf{95.6}&92.7 
&  69.4&  87.6&   91.2& 86.9&  78.8& 90.7& 90.7&91.2&89.1&  83.9&  69.3& 93.4\\ 
         &  AUC
&  \textbf{98.1}&94.8 
&  91.3&  92.1&   94.7& 90.0&  86.5& 95.7& 96.7&92.8&95.9&  84.6&  76.0& 93.6\\ 
         &  $\mathrm{F_{1}}$&  \textbf{94.8}&90.7 
&  85.1&  88.5&   86.7& 85.4&  82.6& 77.4& 78.5&85.6&90.7&  78.8&  71.0& 81.0\\
\hline 
    \end{tabular}

\end{table*}

\noindent\textbf{Performance of auxiliary tasks  of molecular markers and histology prediction.} 
From Table \ref{Performance2}, we observe that M3C2 achieves an AUC of 95.0\%, 90.9\%, 76.0\%, and 98.1\% for IDH mutation, 1p/19q co-deletion, CDKN HOMDEL and NMP prediction, respectively, significantly outperforming all comparison models. Of note, the compared multi-modal learning methods, i.e., MCAT and CMTA, take the molecular markers as input and thus can not perform the molecular markers prediction task.   Figure \ref{fig:roc} presents the ROCs of all models, showcasing the superior performance of our model compared to others.

\noindent\textbf{Generalizability validation.}
We compare our model with SOTA methods on the external validation dataset, i.e., IvYGAP \citep{eberhart2020spatial}, without fine-tuning. Results are shown in the right panel of Table \ref{Performance1}. We can observe from the table that our method achieves the increment of  98.1\%, 99.7\% and 98.1\% over the best SOTA method in accuracy, AUC and $\mathrm{F_{1}}$-score, respectively, suggesting our great robustness and generalizability.

\begin{figure}
    \centering
    \includegraphics[width=1\linewidth]{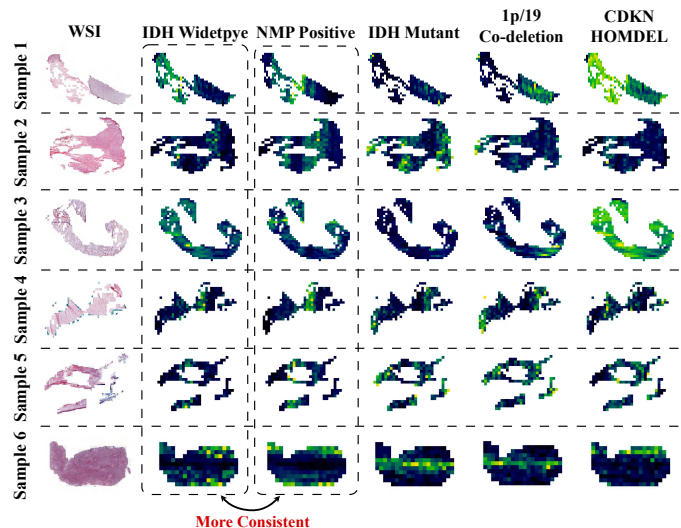}
    \caption{Visualization maps of M3C2 predicting molecular markers of IDH mutation, 1p/19q co-deletion and CDKN HOMDEL, as well as histology of NMP.}
    \label{fig:visualization}
\end{figure}

\noindent\textbf{Network interpretability.}
An additional visualization experiment is conducted based on patch decision scores to test the interpretability of our method.
Figure \ref{fig:visualization} shows the visualization maps of our method predicting molecular markers and histology. As shown in this figure, our method generates more consistent model decision maps for predicting IDH wildtype (molecular marker) with NMP positive (histology feature) compared to other molecular markers. This finding aligns with the current diagnostic criteria where glioblastoma is predominantly IDH wildtype and NMP positive in histology, suggesting our success in modeling cross-modal co-occurrence. Consequently, this indicates the interpretability of integrating molecular markers with histology for clinical diagnosis.

\noindent\textbf{Subgroup analysis.}
To validate the effectiveness of our method on different WSI materials, such as frozen and FFPE sections, we perform additional subgroup analysis on multiple tasks including glioma classification, molecular markers prediction and histology prediction. Specifically, the results are presented  in Table \ref{tab:Subgroup analysis}. As shown in this table,  the performance of frozen sections is slightly inferior, yet still comparable, to that of  FFPE sections for the molecular markers and histology prediction tasks. This is because, compared to frozen sections, FFPE sections preserve tissue architecture and cellular morphology more effectively \citep{ozyoruk2021deep,zhang2022mutual}.
Additionally, on the glioma classification task, the performance is similar between the two materials, with an AUC of 88.1$\%$ for FFPE and 89.2$\%$ for frozen sections, highlighting the practical applicability  of our method in addressing varied clinical needs, especially for different WSI materials.

\begin{table}\small
\centering
\caption{Subgroup analysis in terms of WSI materials on multiple tasks of molecular markers and histology prediction, as well as glioma classification.}
\label{tab:Subgroup analysis}
\begin{tabular}{c||c||ccccc} \toprule\hline  
Material & Task & Acc. & Sen. & Spec. & AUC & F1-score \\ \hline 
\multirow{5}{*}{FFPE} & Diag & 82.1 & 82.1 & 86.4 & 88.1 & 82.1 \\
 & IDH & 90.1& 89.2 & 90.9 & 96.1 & 89.8 
\\ 
 & 1p19q & 80.1 & 96.6 & 69.6 & 92.2 & 79.2 
\\ 
 & CDKN & 80.1 & 76.5 & 83.1 & 82.8 & 77.6 
\\ 
 & Grade & 88.7 & 77.9 & 97.6 & 98.4 & 86.2 
\\ \hline \hline 
\multirow{5}{*}{Frozen} & Diag & 75.2 & 75.2 & 78.7 & 89.2 & 75.2 \\ 
 & IDH & 87.6 & 72.3 & 94.3 & 93.1 & 78.2 \\ 
 & 1p19q & 81.0 & 83.5 & 87.7 & 83.4 & 70.8 \\ 
 & CDKN & 65.4 & 76.1 & 59.2 & 66.6 & 72.5 \\ 
 & Grade & 78.4 & 66.7 & 100.0 & 98.1 & 80.0 \\ \hline\bottomrule

\end{tabular}

\end{table}

\subsection{Results of ablation experiments}
\label{section_ablation}

\begin{table*}\small
\centering

\caption{Ablation study on the multi-modal disentanglement loss, CPLC-Graph network, LC loss and DCC loss on the internal and external validation datasets.}
\label{tab:ablation}
\begin{tabular}{c||ccccc|ccccc} \toprule\hline  
\multirow{2}{*}{Ablation}& \multicolumn{5}{c|}{Internal} & \multicolumn{5}{c}{External} \\ \cline{2-11}
 & Acc. & Sen. & Spec. & AUC & F1-score & Acc. & Sen. & Spec. & AUC & F1-score \\ \hline \hline 
$\mathcal{L}_{\rm disent-v1}$& 65.8 & 65.8 & 85.3 & 86.1 & 65.8 & 94.5 & 94.5 & 57.3 & 99.7 & 94.5 
\\ 
$\mathcal{L}_{\rm disent-v2}$ & 67.8 & 67.8 & 83.1 & 86.1 & 67.8 & 90.4 & 90.4 & 69.3 & 99.1 & 90.4 
\\ 
$w/o$ $\mathcal{L}_{\rm disent}$ & 64.8 & 64.8 & 83.2 & 87.7 & 64.8 & 93.7 & 93.7 & 57.3 & 99.3 & 93.7 
\\ \hline
$w/o$  Graph& 69.4 & 69.4 & 81.3 & 86.4 & 69.4 & 93.1 & 93.1 & \textbf{75.5}& 99.6 & 93.1 
\\  \hline 
$w/o$  LC loss& 62.5 & 62.5 & 86.4 & 86.5 & 62.5 & 90.2 & 90.2 & 93.7 & 98.9 & 90.2 
\\ \hline 
 $w/o$  DCC& 67.4 & 67.4 & \textbf{90.2} & 88.4 & 67.4 & 91.3 & 91.3 & 81.6 & 97.5 &91.3 
\\ \hline\hline
\textbf{Ours}& \textbf{78.6}& \textbf{78.6}& 85.4& \textbf{88.7}& \textbf{78.6}& \textbf{98.1}& \textbf{98.1}& 69.5& \textbf{99.7}& \textbf{98.1}\\ \hline
\bottomrule
\end{tabular}
\end{table*}

\begin{figure}[t]\label{fig:cmg ablation}
    \centering
    \includegraphics[width=.9\linewidth]{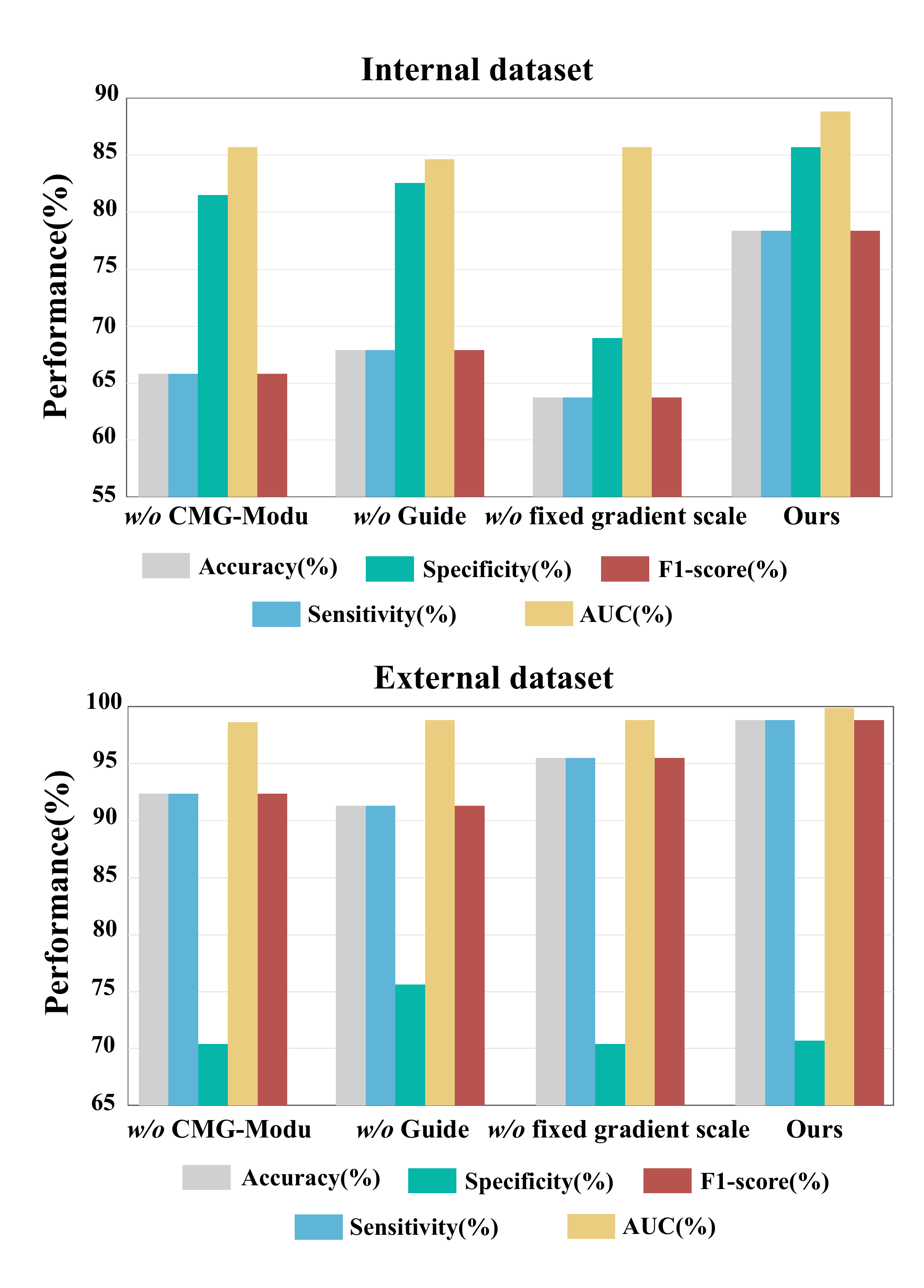}
    \caption{Ablation study of CMG-Modu training strategy on the glioma clasification task over internal and external validation datasets.}
    
\end{figure}

We ablate different components of our M3C2 method to thoroughly analyze their effects on glioma classification and the prediction of molecular markers and histology features.

\noindent\textbf{Multi-scale disentangling.} 
To evaluate the effectiveness of the proposed multi-scale disentangling module, we conduct ablation experiments on the disentanglement loss $\mathcal{L}_{\rm disent}$ as follows: (1) $w/o$ the constraint between the modality independent features of molecular markers ($\mathbf{I}_{\rm m}$) and histology ($\mathbf{I}_{\rm h}$), denoted as $\mathcal{L}_{\rm disent-v1}$; (2) 
$w/o$ the constraint between the independent and shared features within each modality, denoted as $\mathcal{L}_{\rm disent-v2}$; 3) $w/o$ the disentanglement loss. The results on the glioma classification task and the auxiliary tasks are shown in Table \ref{tab:ablation} and Table \ref{Ablation_suppmentary}, respectively. All three models perform worse than M3C2 on both the internal and external validation datasets, suggesting that the designed multi-scale  disentanglement loss can enhance the overall model performance.

% We tested various constraint methods for independent and shared features. Loss α\alpha indicates no constraint on the relationship between the independent features of the two tasks. Loss β\beta signifies no constraint on the relationship between independent and public features within a task. The w/ow/o loss implies no constraints on any task features. Table ?????????????????????????????????????????????\ref{tab:ablation} demonstrates that loss α\alpha exhibits the best overall performance among the three, with an accuracy improvement of 5.0\% compared to the least effective loss β\beta. Other metrics also show relative improvements. However, there remains a substantial performance gap compared to the optimal setting that retains all constraints. Similar trends are observed in external datasets, underscoring the efficacy of the proposed Multi-Scale Disentangling method.

\noindent\textbf{CPLC-Graph network.}
Table \ref{tab:ablation} shows that, by setting graph balancing weight $\alpha$ to 0 for the proposed CPLC-Graph, the accuracy, sensitivity, specificity, AUC and $\mathrm{F_{1}}$-score decreases by 9.2\%, 9.2\%, 4.1\%, 2.3\% and 9.2\%, respectively. Similar results are observed for the prediction tasks of molecular markers and histology (Table \ref{Ablation_suppmentary}). Also, the ROC of removing the CPLC-Graph network is shown in Figure \ref{fig:roc}. These results indicate the utility of the proposed CPLC-Graph network. 

\noindent\textbf{LC loss.}
We further evaluate the effectiveness of our LC loss, by simply removing the LC loss when training our M3C2 method. 
Table \ref{tab:ablation} shows that the performance after removing LC loss decreases in four main metrics, causing a reduction of 16.1\%, 16.1\%, 2.2\% and 16.1\%, in accuracy, sensitivity, AUC and $\mathrm{F_{1}}$-score, respectively. Similar results for the tasks of molecular marker and histology prediction are observed in the Table \ref{Ablation_suppmentary} with the ROC in Figure \ref{fig:roc}, indicating the effectiveness of the LC loss.

\noindent\textbf{DCC loss.}
From Table \ref{tab:ablation}, we observe that the proposed DCC loss improves the performance in terms of accuracy by 11.2\%. Similar results can be found for sensitivity, AUC and $\mathrm{F_{1}}$-score. From Table \ref{Ablation_suppmentary}, we observe that the AUC decreases 8.9\%, 2.6\%, 0.9\% and 5.1\% for the prediction of IDH, 1p/19q, CDKN and NMP, respectively, when removing the DCC loss. Such performance is also found in comparing the ROC curves in Figure \ref{fig:roc}, suggesting the importance of the DCC loss for all the tasks. 

\noindent\textbf{CMG-Modu training strategy.}
Finally, we validate the effectiveness of the CMG-Modu training strategy, by modifying our training strategy as follows:
(1) $w/o$ fixed gradient scale - directly project the modulated gradient to the perpendicular direction of the dominant gradient without scaling; (2) $w/o$ guide - remove the guidance of histology grade during gradient modulation; (3) $w/o$ CMG-Modu - remove the whole CMG-Modu training strategy. The experimental results are shown in Figure \ref{fig:cmg ablation},  in which all three modified training strategies perform worse than M3C2, indicating the effectiveness of our CMG-Modu training strategy. 

Overall, as shown in the left panel of Table \ref{tab:ablation}, the glioma classification performance of our M3C2 slightly decreases when ablating each individual module. For example, the AUC value  decreases by only 0.3\% and 1.0\% when ablating the dynamic confidence constrain (DCC) loss and disentanglement loss on the internal validation dataset, respectively, indicating the relative independence of these modules for glioma classification. Similar results can also be observed on the external validation dataset, as shown in the right panel of Table \ref{tab:ablation}.

Moreover, in Table \ref{tab:ablation}, the superior performance observed on the external dataset compared to the internal dataset is primarily attributed to differences in their class distributions, i.e., the external dataset consists exclusively of two classes: grade 4 GBM and high-grade astrocytoma. The reduced number of classes in the external dataset contributes to the relatively higher performance.
To ensure a fair comparison, we have recalculated all experimental results using only the two classes (grade 4 GBM and high-grade astrocytoma) from the internal dataset. These results are provided in Supplementary Table 3. As shown, the model's performance on the external dataset is comparable to its performance on the internal dataset, further demonstrating the generalizability of our model.

\begin{table*}[t]
\scriptsize
    \centering
    \caption{Ablation study on the multi-modal disentanglement loss, CPLC-Graph network, LC loss, DCC loss and CMG-Modu training strategy on the auxiliary tasks of molecular markers and histology prediction over internal dataset.}
    \label{Ablation_suppmentary}
    \begin{tabular}{|c|c|clcccccccc|} \hline 
         Task&  Metrics&  Ours 
& $\mathcal{L}_{\rm disent-v1}$
&  $\mathcal{L}_{\rm disent-v2}$ 
&  $w/o$ $\mathcal{L}_{\rm disent}$ 
&   $w/o$  Graph
& $w/o$  LC loss
&  $w/o$  DCC
& $w/o$  GMC-Modu
& $w/o$  Guide&$w/o$   fixed gradient scale\\ \hline 
         \multirow{5}{*}{\rotatebox{90}{IDH}}&  Acc.&  \textbf{88.8}
&80.6 &  90.4 &  80.9 &   81.6 & 82.6 &  78.9 & 83.2 & 82.6 &85.5 
\\  
         &  Sen.&  
82.6  
&\textbf{93.4}&  68.8 &  81.0 &   81.0 & 82.6 &  71.1 & 80.2 & 87.6 &89.3 
\\  
         &  Spec.&  \textbf{92.9}
&72.1 &  90.9 &  80.9 &   82.0 & 82.5 &  84.2 & 85.2 & 79.2 &83.1 
\\ 
         &  AUC&  
\textbf{95.0}
&89.0 &  92.7 &  87.8 &   87.2 & 90.0 &  86.1 & 90.5 & 88.2 &93.5 
\\ 
         &  $\mathrm{F_{1}}$-score&  \textbf{85.5}
&79.3 &  26.8 &  77.2 &   77.8 & 79.1 &  72.9 & 79.2 & 80.0 &83.1 
\\ \hline 
 \multirow{5}{*}{\rotatebox{90}{1p19q}}& Acc.
& 
80.6  
&74.3 & 73.4 & 76.0 &  73.7 & 75.7 &  \textbf{80.9}& 78.3 & 74.0 &79.9 
\\ 
 & Sen.
& 81.7&82.9 & \textbf{84.1}& 51.2 &  72.0 & 85.4 &  84.1 & 57.3 & 74.4 &59.8 
\\ 
 & Spec.
& 
80.2  
&71.2 & 69.4 & 85.1 &  74.3 & 72.1 &  79.7 & 86.0& 73.9 &\textbf{87.4} 
\\ 
 & AUC
& \textbf{90.9}
&83.1 & 81.6 & 81.9 &  82.5 & 83.0 &  88.3 & 84.3 & 84.2 &86.3 
\\  
 & $\mathrm{F_{1}}$-score& 
69.4 
&63.6 & 63.0 & 53.5 &  59.6 & 65.4 &  \textbf{70.4}& 58.7 & 60.7 &61.6 
\\ \hline
 \multirow{5}{*}{\rotatebox{90}{CDKN}}& Acc.
&  
\textbf{72.7}
&68.8 & 61.2 & 66.8 &  68.4 & 68.1 
&  67.4 
& 70.4 & 69.4 &72.0 
\\ 
 & Sen.
& 76.3  
&71.3 & \textbf{95.0}& 61.3 &  75.0 & 63.1 
&  91.9 
& 81.3 & 75.0 &78.1 
\\  
 & Spec.
& 
68.8  
&66.0 & 23.6 & 72.9 &  61.1 & \textbf{73.6}&  40.3 
& 58.3 & 63.2 &65.3\\  
 & AUC
& \textbf{76.0}
&72.8 & 72.7 & 75.5 &  71.8 & 75.2 
&  75.1 
& 73.8 & 74.6 &73.8 
\\ 
 & $\mathrm{F_{1}}$-score& 
\textbf{74.6}&70.6 & 72.0 & 66.0 &  71.4 & 67.6 
&  74.8 
& 74.3 & 72.1 &74.6 
\\ \hline 
 \multirow{5}{*}{\rotatebox{90}{NMP}}& Acc.
&   \textbf{94.4}
&84.2 & 87.8 & 88.2&  87.2 & 92.1 &  84.9& 90.1 & 87.5 &91.1 
\\ 
 & Sen.
&

\textbf{93.4}&82.0 & 85.6 & 83.2 &  94.0 & 93.4 &  92.2 & 88.6 & 83.2 &91.0 
\\ 
         &  Spec.
&  

\textbf{95.6}
&86.9 &  90.5 &  94.2 &   78.8 & 90.5 &  75.9 & 92.0 & 92.7 &91.2 
\\ 
         &  AUC
&  \textbf{98.1}
&93.1 &  93.3 &  93.8 &   93.1 & 96.3 &  93.0 & 94.7 & 93.5 &96.2 
\\ 
         &  $\mathrm{F_{1}}$-score&  
\textbf{94.8}&85.1 &  88.5 &  88.5 &   89.0 & 92.9 &  87.0 & 90.8 & 88.0 &91.8 
\\
\hline 
    \end{tabular}
\end{table*}

\subsection{Analysis on experimental settings}
\noindent\textbf{Influence of input magnification.}
We evaluate the performance of our M3C2 method on the glioma diagnosis task under different  input magnification settings, i.e., using only one magnification of 10X and 20X, as well as simply concatenating and adding features of the two magnifications. Table \ref{tab:input magnification} tabulates the results of glioma classification under different magnification settings. As shown in this table, the glioma prediction performance is improved using our proposed multi-scale disentangling module, compared to other settings of single magnification or multi-magnification concatenation and addition. For instance, the classification results of our method are 78.6$\%$, 78.6$\%$, 85.4$\%$ , 88.7\% and 78.6$\%$ in accuracy, sensitivity, specificity, AUC and F1-score,  whereas those of multi-magnification concatenation are 69.7$\%$, 69.7$\%$, 81.6$\%$, 88.6$\%$ and 69.7$\%$, respectively. To summarize, the above results imply the performance improvement of our method in efficient multi-magnification fusion.

Moreover, additional experimental results on the auxiliary tasks of molecular markers and histology prediction with different input magnification settings can be found in Table 2 in the supplementary material.

\begin{table}\small
%\scriptsize
\centering
\caption{Mean values in terms of percentage for glioma classification accuracy by our and other magnification-related baseline methods.}
\label{tab:input magnification}
\begin{tabular}{c|ccccc} \toprule  
\multicolumn{6}{c}{Internal dataset} \\ \hline 
Scale & Acc. & Sen. & Spec. & AUC & F1-score \\ \hline 
10X & 68.8 & 68.8 & \textbf{88.7}& 88.4 & 68.8 
\\  
20X & 66.4 & 66.4 & 88.4 & 88.5 & 66.4 
\\ 
10X+20X (concat)& 69.7 & 69.7 & 81.6 & 88.6 & 69.7 
\\ 
 10X+20X (add)& 74.3 & 74.3 & 74.1 & 88.6 &74.3 
\\
 \textbf{Ours}& \textbf{78.6}& \textbf{78.6}& 85.4& \textbf{88.7}&\textbf{78.6}\\\hline \hline 
\multicolumn{6}{c}{External dataset} \\ \hline 
Scale & Acc. & Sen. & Spec. & AUC & F1-score \\  \hline 
10X & 95.8 & 95.8 & 39.1 & 99.6 & 95.8 
\\ 
20X & 93.1 & 93.1 & 63.4 & 99.3 & 93.1 
\\ 
10X+20X (concat) & 97.3 & 97.3 & 39.1 & 99.7 & 97.3 
\\ 
 10X+20X (add)& 96.5 & 96.5 & 57.3 & 99.6 &96.5 
\\ 
 \textbf{Ours}& \textbf{98.1}& \textbf{98.1}& \textbf{69.5}& \textbf{99.7}&\textbf{98.1}\\\bottomrule

\end{tabular}
\end{table}

\noindent\textbf{Impact of auxiliary tasks for glioma classification.}
In our method, we perform multi-task learning with the primary task of glioma classification and the auxiliary tasks of molecular and histology prediction. Thus, we conduct additional experiments to evaluate the effectiveness of the involved auxiliary tasks on the glioma classification performance, via ablating the  molecular and/or histology prediction tasks. As shown in Table \ref{tab:auxiliary}, the glioma classification accuracy decreases 9.5$\%$, 10.5$\%$ and 10.8$\%$ on the internal dataset when training without histology prediction module, molecular prediction module and both the two modules, respectively. Similar results can be found on the external validation dataset. This indicates that our multi-task learning strategy can facilitate the glioma classification performance via incorporating the two  auxiliary tasks.

\begin{table}\small
\centering
\caption{Mean values in terms of percentage for glioma classification accuracy by our method with and without auxiliary tasks.}
\label{tab:auxiliary}
\begin{tabular}{c|ccccc} \toprule
\multicolumn{6}{c}{Internal dataset} \\ \hline 
Structure & Acc. & Sen. & Spec. & AUC & F1-score \\ \hline 
$w/o$ histology & 69.1 & 69.1 & 80.6 & 87.5 & 69.1 \\ 
$w/o$ markers & 68.1 & 68.1 & 74.3 & 88.3 & 68.1 \\
 $w/o$  both tasks & 67.8 & 67.8 & 70.2 & 87.1 &67.8 \\
 \textbf{Ours}& \textbf{78.6}& \textbf{78.6}& \textbf{85.4}& \textbf{88.7}&\textbf{78.6}\\\hline\hline 
\multicolumn{6}{c}{External dataset} \\ \hline 
Structure & Acc. & Sen. & Spec. & AUC & F1-score \\ \hline 
$w/o$ histology & 93.4 & 93.4 & 69.5 & 99.5 & 93.4 
\\ 
$w/o$ markers & 92.1 & 92.1 & \textbf{87.7}& 99.6 & 92.1 
\\ 
 $w/o$ both tasks & 92.6 & 92.6 & 57.2 & 99.3 &92.6 
\\ 
 \textbf{Ours}& \textbf{98.1}& \textbf{98.1}& 69.5& \textbf{99.7}&\textbf{98.1}\\\bottomrule
\end{tabular}
\end{table}

\section{Discussion}

%第一段思路：clinical motivation and related work limitation

Recently, cancer diagnosis criteria has shifted from solely relying on histology features to integrating molecular markers and histology features, especially for brain cancers such as diffuse glioma. 
%However, obtaining the molecular markers is both expensive and time consuming, usually costing weeks to months for the genomic test. Hence, 
It is promising to develop DL-based methods to facilitate clinical diagnosis of 
glioma. Some recent works, such as Charm \citep{charm} and Deepglioma \citep{khalighi2024artificial}, have been specially designed for glioma classification based on WHO 2021 criteria. Yet, these existing methods suffer from either incomplete diagnosis pipeline or inability of cross-modal modelling of molecular and histology features. In this context, multi-modal learning (MML) \citep{zhang2024prototypical} seems to be a potential solution for the integrated diagnosis of glioma. Traditional MML methods take molecular markers as input, which are not predicted from the WSIs, making them less practical in real-world applications.

%第2段思路： highlight our cross modal interaction / multi scale

In contrast, we propose a multi-task learning method (M3C2) to jointly predict histology features, molecular markers, and glioma classes, incorporating efficient task interaction designs, including the DCC loss and CMG-Modu learning strategy. 
The DCC loss constrains the histology features of NMP and molecular features of IDH to be close, based on the clinical observation that over 95$\%$ of patients with NMP are IDH wildtype \citep{alzial2022wild}.  many  histology-molecular pairs exhibit strong correlations beyond the widely recognized associations between IDH wild-type and NMP. For example, the 1p/19q co-deletion is strongly associated with the characteristic histological features of oligodendroglioma, such as the \textit{fried egg} cellular appearance\citep{wesseling2015oligodendroglioma}.
Similarly, microvascular proliferation and H3 K27me3 loss are associated pairs in ependymomas. Beyond gliomas, many cancers also exhibit histology-molecular correlations. For instance, in lung mucinous adenocarcinoma, the presence of mucinous histology strongly correlates with KRAS mutations, representing another critical histology-molecular relationship frequently used in clinical practice..
Therefore, the proposed DCC loss is not limited to this specific pair and can be generalized  not only to other glioma subtypes but also to various cancers, where such histology-molecular correlations play a critical role in diagnosis.
 Moreover, the CMG-Modu allows gradient guidance from the decisive modality, either histology features or molecular markers, during training. In this strategy, histology prediction serves as the reference for molecular prediction when the WSI shows NMP features, and vice versa. This is motivated by the fact that, under the revised criteria, patients with NMP are still considered as high grade, whereas the grade of patients without NMP features is determined by molecular markers. Furthermore, ablation study results (Table \ref{tab:ablation} and Figure \ref{fig:cmg ablation}) validate the effectiveness of our task interaction design.

In addition to the task interaction design, we also propose a multi-scale disentangling module for extracting multi-magnification histology features, which can be applied broadly to pathological image analysis for various diseases. This is particularly relevant in clinical settings, where multi-magnification histopathology images play a vital role in precision medicine, as each magnification level provides unique diagnostic insights. Specifically, high-magnification images capture fine-grainedcellular-level features, such as intricate cell morphology, nuclear atypia, and mitotic activity, which are essential for detailed cellular characterization. Low-magnification images capture tissue-level features, providing a holistic view of  tissue architecture, structural organization, and spatial relationships among tissue niches. 
These cellular-level and tissue-level  features offer complementary perspectives and enable a comprehensive understanding of disease, enhancing diagnostic accuracy. By integrating both scales, our framework supports robust and clinically meaningful insights, making it broadly applicable to various types of cancer and other diseases that require detailed histopathological evaluation. To validate the impact of our multi-magnification approach, we conduct experiments (Table \ref{tab:input magnification}) comparing our method to single magnification and simple combination methods. The significant performance improvement demonstrates the effectiveness of our multi-magnification design.

To further explore the superiority of each modality on specific groups of patients, we divide the patients into four groups based on their glioma classes. For each group, we generate a histogram where the y-axis reflects the superiority of each modality, defined as the ratio of the importance of histology features to molecular markers in the final glioma grading task. Specifically, the importance value of each modality in the final glioma grading task is derived from the gradient scale of the histology and molecular prediction task in our cross-modal gradient modulation strategy. The histograms of the four groups are shown in Figure 3 in supplementary material. As demonstrated, the superiority of each modality varied across groups, which could provide more interpretability for the joint model and suggest significant  potential for personalized treatment in the future.

Our method has great potential in real-world  clinical scenarios, including the computer-assisted diagnosis of glioma and prediction of crucial molecular markers. 
The blueprint of the clinical application of our M3C2 method is illustrated in Figure 2 of the supplementary material.
As shown, compared to the current clinical glioma diagnosis process
, our proposed AI system (M3C2) can make diagnosis with less expenses, especially in obtaining molecular markers.
In addition, our method can be further deployed in hospitals to offer real-time, AI-assisted glioma diagnosis based on WSIs, potentially improving the diagnosing efficiency and further accelerating the treatment planning for patients. 
Moreover, our M3C2 method can be used in many other clinical scenarios, such as the rapid diagnosis during neuro-therapy with frozen sections, thereby increasing the rate of successful surgeries.
Overall, these potential real-world applications imply how our M3C2 can contribute to precision neuro-oncology.

% weakness: only focus on markers essential for diagnosis, yet more is need for treatment plan; due to the difference of clicnail resource, is is hard to guarentee modality completeness, especially for retrospective cases,  e.g., some patient do not have CDKN. We should have proposed a method tailored for this scenario, e.g,. modality-imputaion, weakly supervise learnig

Despite the high performance of classifying glioma, our M3C2 still has some weakness.
For example, we only pay attention to predicting molecular markers essential for glioma diagnosis, without considering other molecular markers important for treatment planning and disease prognosis, e.g., MGMT (O-6-methylguanine-DNA methyltransferase) methylation.
Furthermore, our method is dependent on the completeness of the multi-modal data of histology and molecular markers. However, certain modalities, e.g., molecular markers, could be missing in real-world scenarios. For instance, the molecular markers of CDKN might be inaccessible for patients with NMP features, especially for those from retrospective study.  
One possible solution for incomplete modalities is to propose a more modal-adaptable approach, such as incorporating modality-imputation techniques or leveraging weakly supervised learning algorithms. 

% future work: more potential markers,MGMT EGFR ATRX; more magnifications; other cancers

In future work, it is essential to explore additional potential biomarkers, such as 
% MGMT: O-6-methylguanine-DNA methyltransferase
% EGFR: Epidermal growth factor receptor
% ATRX: Alpha thalassemia/mental retardation syndrome X-linked
MGMT methylation, EGFR (epidermal growth factor receptor) amplification, ATRX (alpha thalassemia/mental retardation syndrome X-linked) mutation, to enhance the diagnostic and prognostic capabilities of our study. These markers could provide more comprehensive insights into tumour behavior and patient treatment outcomes. Furthermore, using more WSI magnifications, e.g., 40X and 5X, can also promise to facilitate the efficient extraction of morphology feature for better model performance. 
Finally, according to the most recent WHO cancer diagnostic criteria, the diagnosis of many other cancers, including endometrial cancer \citep{imboden2021implementation}, renal neoplasia \citep{trpkov2021new} and thyroid carcinomas \citep{volante2021molecular}, has also shifted from purely relying on histology features to integrating molecular and histology features. Hence, our proposed M3C2 for cancer classification framework can be applicable for these cancers by jointly predicting molecular markers and histology features while modelling their interactions for cancer classification.

\section{Conclusion}
The paradigm of pathology diagnosis of diffuse gliomas has shifted to integrating molecular makers with histology features. In this paper, we target on classifying glioma under the latest diagnosis criteria, via jointly learning the tasks of molecular marker and histology prediction, as well as the final glioma classification.
Inputting multi-magnification histology WSIs, our model incorporates a novel AHMT-MIL framework with multi-scale disentangling to extract both cellular-level and tissue-level  information for the downstream tasks. Moreover, a CPLC-Graph network is devised for intra-omic interactions, while a DCC loss and a CMG-Modu training strategy are further designed for inter-omic interactions. 
Our experiments demonstrate that M3C2 achieves superior and more robust performance over other state-of-the-art methods, opening a new avenue of for digital pathology based on WSIs in the era of molecular pathology. 

\section{CRediT authorship contribution statement}
\textbf{Xiaofei Wang}: Writing – original draft, Visualization, Validation, Methodology, Investigation, Formal analysis, Data curation, Conceptualization.
\textbf{Hanyu Liu}: Writing – original draft, Visualization, Validation, Methodology, Investigation, Formal analysis, Data curation.
\textbf{Yupei Zhang}: Writing – original draft,  Validation, Methodology, Conceptualization.
\textbf{Boyang Zhao}: Writing – review \& editing, Visualization, Validation, Data curation.
\textbf{Stephen Price}: Resources, Funding acquisition.
\textbf{Chao Li}: Conceptualization, Validation, Supervision, Resources, Funding acquisition, Methodology, Writing – review \& editing, Project administration.

\section{Data and code availability}
This paper uses three publicly available datasets, including TCGA GBM-LGG dataset (\href{https://portal.gdc.cancer.gov/}{https://portal.gdc.cancer.gov/}), CPTAC dataset (\href{https://portal.gdc.cancer.gov/}{https://portal.gdc.cancer.gov/}) and IvYGAP dataset (\href{https://glioblastoma.alleninstitute.org/static/home/}{https://glioblastoma.alleninstitute.org/static/home/}). The source code is available at \href{https://github.com/LHY1007/M3C2/}{https://github.com/LHY1007/M3C2/}

\section{Declaration of generative AI and AI-assisted technologies in the writing process}
The authors declare that they have nothing to disclose.

\section{Declaration of competing interest}
Xiaofei Wang reports financial support was provided by China Scholarship Council. Stephen Price reports financial support was provided by National Institute for Health and Care Research. Chao Li reports financial support was provided by Guarantors Of Brain.

\section{Acknowledgements}
This work was supported by the NIHR Brain Injury MedTech Co-operative and the NIHR Cambridge Biomedical Research Centre (NIHR203312) and Addenbrooke’s Charitable Trust. This work presents independent research funded by the National Institute for Health and Care Research (NIHR). The views expressed are those of the author(s) and not necessarily those of the NHS, the NIHR or the Department of Health and Social Care.

% \section*{References}

% \section*{\itshape Reference style}
% Text: All citations in the text should refer to:
% \begin{enumerate}
% \item Single author: the author's name (without initials, unless there
% is ambiguity) and the year of publication;
% \item Two authors: both authors' names and the year of publication;
% \item Three or more authors: first author's name followed by `et al.'
% and the year of publication.
% \end{enumerate}
% Citations may be made directly (or parenthetically). Groups of
% references should be listed first alphabetically, then chronologically.

%%Harvard
\bibliographystyle{model2-names.bst}\biboptions{authoryear}
\bibliography{main_revised_unmarked}

% \section*{Supplementary Material}

\end{document}